\PassOptionsToPackage{numbers}{natbib}
\PassOptionsToPackage{sort}{natbib}

\documentclass{article}

\usepackage{geometry}
\usepackage[utf8]{inputenc} 
\usepackage[T1]{fontenc}   
\usepackage{fullpage}
\usepackage[numbers]{natbib}
\newcommand{\comment}[1]{}
\date{}

\usepackage{microtype}
\usepackage{graphicx}
\usepackage{subcaption}
\usepackage{booktabs} 
\usepackage{floatrow}

\usepackage{graphicx}
\usepackage{amsmath}
\usepackage{amssymb}

\usepackage[pagebackref=true,breaklinks=true,letterpaper=true,colorlinks,bookmarks=false]{hyperref}
\usepackage{tablefootnote}
\usepackage{amsmath}
\usepackage{amssymb}
\usepackage{tabularx}
\usepackage{arydshln}
\usepackage[T1]{fontenc}    
\usepackage{url}            
\usepackage{booktabs}       
\usepackage{amsfonts}       
\usepackage{nicefrac}       
\usepackage{microtype}      
\usepackage{algorithm}
\usepackage{algorithmic}
\usepackage{lipsum}
\usepackage{enumitem}
\usepackage{amsmath, amssymb, amsthm}
\usepackage{wrapfig}
\usepackage{arydshln}
\usepackage[utf8]{inputenc}
\usepackage{mathtools}
\usepackage{subcaption}
\usepackage{caption}
\usepackage{bbm}
\usepackage{array}
\usepackage{arydshln}

\usepackage{multirow}
\usepackage{url}            
\usepackage{booktabs}       
\usepackage{amsfonts}       
\usepackage{nicefrac}       
\usepackage{xcolor}
\usepackage{microtype}      
\usepackage{amssymb,amsmath, amsthm}
\usepackage[utf8]{inputenc}
\usepackage{mathtools}
\usepackage{subcaption}
\usepackage{tikz}
\usetikzlibrary{calc}
\usepackage{graphicx}
\usepackage{caption}
\usepackage{bbm}
\usepackage{bm}
\usepackage{array}
\usepackage{arydshln}
\usepackage{xspace}

\usepackage{algorithmic}
\usepackage[colorinlistoftodos,prependcaption]{todonotes}

\definecolor{gr}{rgb}{0.25, 0.25, 0.25}

\newcommand{\fedl}{\textsc{FedLabel}}

\newcommand{\indi}{\mathbbm{1}}

\newcommand{\ec}{M}

\newcommand{\ic}{k}

\newcommand{\dtoprule}{\specialrule{1pt}{0pt}{0.7pt}%
            \specialrule{0.3pt}{0pt}{\belowrulesep}%
            }
\newcommand{\dbottomrule}{\specialrule{0.3pt}{0pt}{0.7pt}%
            \specialrule{1pt}{0pt}{0.1pt}%
            }
            
\newcommand{\gf}{F}

\newcommand{\lff}{f}

\newcommand{\dunl}{\mathcal{D}_{U,k}}
\newcommand{\sbl}{\mathbf{s}}

\newcommand{\psela}{\hat{y}}

\newcommand{\wb}{\mathbf{w}}
\newcommand{\xb}{\mathbf{x}}

\newcommand{\rdat}{p_k}

\newcommand{\defeq}{\vcentcolon=}

\newcommand{\fsup}{F_{\mathcal{L},k}}
\newcommand{\fun}{F_{\mathcal{U},k}}

\newcommand{\lr}{\eta}

\newcolumntype{?}{!{\vrule width 1pt}}

\DeclareMathOperator*{\argmax}{arg\,max}
\DeclareMathOperator*{\argmin}{arg\,min}

\newcommand{\stl}{\mathcal{S}^{(t)}(\xi)}

\usepackage{cleveref}
\crefname{equation}{}{}
\Crefname{equation}{}{}
\Crefname{figure}{Fig.}{Figs.}

\begin{document}


\title{\textbf{Local or Global: Selective Knowledge Assimilation\\ for Federated Learning with Limited Labels}
}

\author{Yae Jee Cho\dag \\
\small Carnegie Mellon University\\
\small \texttt{\href{mailto:yaejeec@andrew.cmu.edu}{yaejeec@andrew.cmu.edu}} 
\and Gauri Joshi \\
\small Carnegie Mellon University\\
\small \texttt{\href{mailto:gaurij@andrew.cmu.edu}{gaurij@andrew.cmu.edu}}\\
\and Dimitrios Dimitriadis\dag \\
\small Amazon\\
\small \texttt{\href{mailto:}{dbdim@amazon.com}}
}
\maketitle


\begin{abstract}
Many existing FL methods assume clients with fully-labeled data, while in realistic settings, clients have limited labels due to the expensive and laborious process of labeling. Limited labeled local data of the clients often leads to their local model having poor generalization abilities to their larger unlabeled local data, such as having class-distribution mismatch with the unlabeled data. As a result, clients may instead look to benefit from the global model trained across clients to leverage their unlabeled data, but this also becomes difficult due to data heterogeneity across clients. In our work, we propose \fedl~where clients selectively choose the local or global model to pseudo-label their unlabeled data depending on which is more of an expert of the data. We further utilize both the local and global models' knowledge via global-local consistency regularization which minimizes the divergence between the two models' outputs when they have identical pseudo-labels for the unlabeled data. Unlike other semi-supervised FL baselines, our method does not require additional experts other than the local or global model, nor require additional parameters to be communicated. We also do not assume any server-labeled data or fully labeled clients. For both cross-device and cross-silo settings, we show that \fedl~outperforms other semi-supervised FL baselines by $8$-$24\%$, and even outperforms standard fully supervised FL baselines ($100\%$ labeled data) with only $5$-$20\%$ of labeled data.

\end{abstract}

\def\thefootnote{\dag}\footnotetext{Work done while at Microsot Research.}\def\thefootnote{\arabic{footnote}}

\vspace{-1em}
\section{Introduction} 
\begin{figure}[!t]
\begin{floatrow}
\ffigbox{\includegraphics[width=0.5\textwidth]{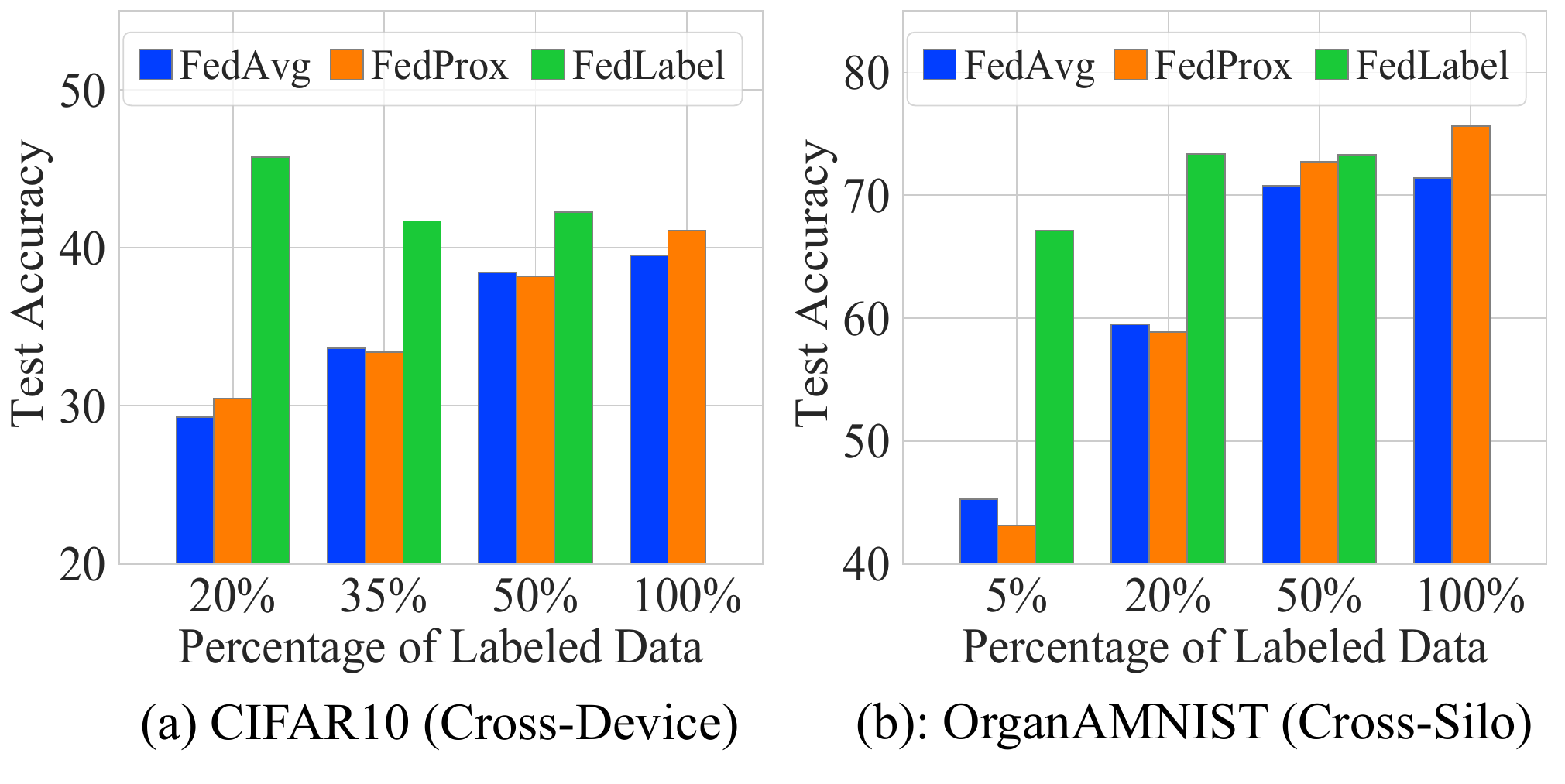}}
{\caption{ Test accuracy of the global model for varying amount of labeled local data in each client. The fewer the labeled data, the lower the test accuracy for standard FL algorithms (FedAvg, FedProx) while our proposed \fedl~performs significantly higher. For CIFAR10, \fedl~achieves an even higher test accuracy than when $100\%$ of labels are used.} \label{fig:moti} } 
\capbtabbox{\renewcommand{\arraystretch}{1.2} 
\centering 
\setlength\tabcolsep{3pt}\small
\begin{tabular}{@{}l||c|c@{}}  
\dtoprule 
\multicolumn{1}{c||}{High Data}   & $20\%$ Labeled & $50\%$ Labeled  \\ 
\multicolumn{1}{c||}{Heterogeneity} & Data & Data \\ \hline
Only Local & $32.57~{\scriptstyle (\pm 2.20)}$ & $41.21~{\scriptstyle (\pm 1.80)}$ \\  
Only Global & $37.51~{\scriptstyle (\pm 1.80)}$  & $38.43~{\scriptstyle (\pm 0.94)}$ \\  \hdashline
Global+Local (\textbf{ours}) & $\mathbf{44.51}~{\scriptstyle (\pm 1.85)}$ & $\mathbf{50.53}~{\scriptstyle (\pm 1.74)}$ \\
 \dbottomrule   
\end{tabular}}{\vspace{2em}\caption{ Test acc. (CIFAR10) when only the local or global model is used for pseudo-labeling which largely underperforms the case when both models are selectively used by our proposed \fedl.  \label{tab:locgloabl}}}
\end{floatrow}
\end{figure}

\vspace{-0.2em}
Federated learning (FL)~\cite{mcmahan2017communication} enables collaborative learning across clients without explicit disclosure of their local data~\cite{kairouz2019advances,wang2021field}. In FL, a server updates its global model by aggregating the local gradients obtained from clients' training on their datasets. These clients can be a number of edge-devices such as  cell-phones (cross-device)~\cite{yang2022crodevice} or a handful of hospitals, for example, willing to train a model for disease prediction without sharing patients' private data (cross-silo)~\cite{sad2020crosilo}. While FL can indeed allow clients to train a single global model without private data sharing, a crucial yet often overlooked limitation found in realistic FL scenarios is that labels can be scarce~\cite{he2021ssfl,fan2022ssfl,che2021ssflnew,kim2022ssflnew,scott2022ssflnew}. In cross-device settings, owners of edge-devices rarely go through the effort of labeling all of their local data such as photos, resulting in only a few labeled samples and a large volume of unlabeled data. Similarly in cross-silo settings, such as hospitals collaborating for predicting diseases, labeling is often a laborious process where healthcare experts are required to process volumes of patients' data~\cite{droz2022medi,zhong2020medcovid19}. Such scarcity of labels can lead to severe performance degadation as shown in \Cref{fig:moti}.

\newfloatcommand{capbtabbox}{table}[][0.47\textwidth]

\begin{wrapfigure}{r}{0.5\textwidth} 
\centering \vspace{-0.1em}
\includegraphics[width=1\textwidth]{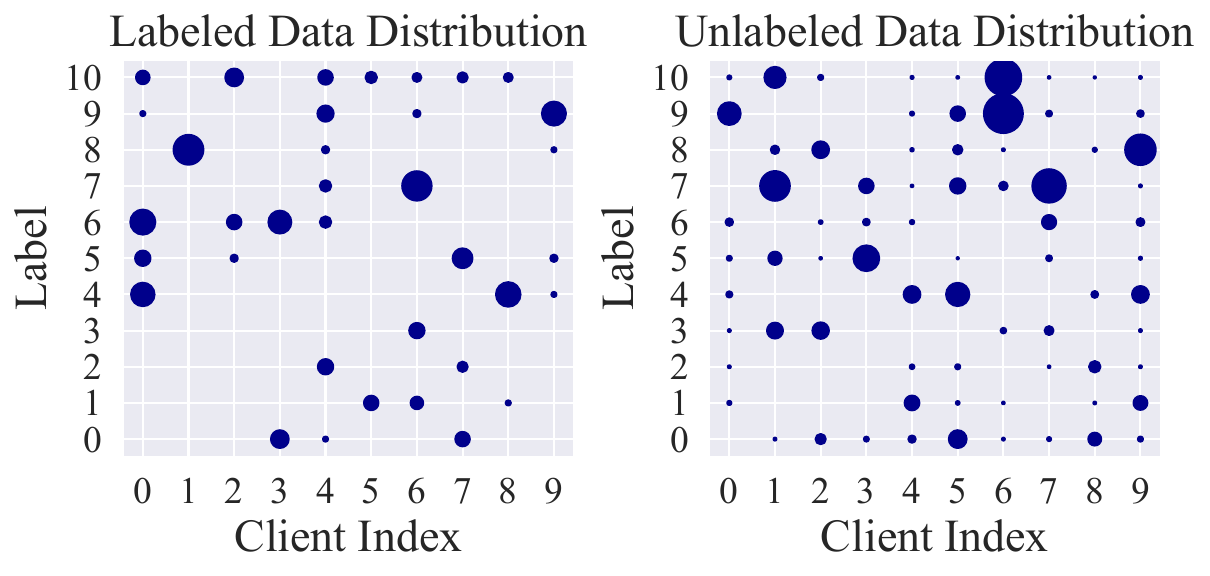} 
 \vspace{-1em} \caption{Class distribution mismatch (CIFAR10) is shown just by limiting the number of labeled data to $20\%$ without artificially biasing the labeled data distribution.\vspace{-1.6em}} \label{fig:moti2}   
\end{wrapfigure}
A na\"ive approach to tackle label scarcity in clients is using standard semi-supervised learning (SSL) methods devised for general machine learning (ML) applications at each client, using its own local data. For example, consistency regularization methods~\cite{xie2020uda} or pseudo-labeling methods~\cite{sohn2020fixmatch} can be directly used with each client's local data with its local model. However, with limited labels, the feature-label pair distribution of the labeled data
 can be different from that of the unlabeled data as shown in \Cref{fig:moti2} and previous work~\cite{zhao2022dcssl,chen2020ssl}. We call this difference between the two distributions as \textit{class distribution mismatch}. As such, the limited labeled data can poorly generalize to a large number of unlabeled local data. This is also shown in \Cref{tab:locgloabl}, where the `Only Local' performance largely degrades for a smaller number of labeled data. Due to these
 scenarios, \textit{leveraging only the local knowledge of a client may not be enough to fully utilize its unlabeled data}. 
Thus, in FL, a client can seek to utilize the global knowledge shared across clients through the aggregation of the local updates to the global model. The problem, however, with only using the global knowledge for leveraging the unlabeled data at the clients is that the data distribution amongst clients' local data is heterogeneous~\cite{sahu2019federated,reddi2020adaptive, haddadpour2019convergence,malinovsky2020local, khaled2020tighter, wang2020tackling}. Due to this \textit{data heterogeneity}, for clients whose local data distribution differs from the overall global data distribution, the global model may also not be useful in assisting the clients to leverage their unlabeled data as shown in \Cref{tab:locgloabl}, `Only Global' case in which the test accuracy is lower than selectively leveraging both global and local models.  

Based on the observations above, either the local or global model, or both, can be useful for clients to leverage their unlabeled data depending on the labeled data's generalizability to the unlabeled data and clients' data heterogeneity. Therefore, to utilize the setting of FL where clients have access to both the knowledge from their local data and the global model, we propose a selective knowledge assimilation method named \fedl~where each client chooses between its local and global model to pseudo-label its unlabeled data based on each model's confidence score. Moreover, with our proposed global-local consistency regularization, we fully utilize both the local and global models when both have useful knowledge of the unlabeled data.

Most relevant line of recent work to \fedl~has proposed the server to identify multiple experts for each client. The experts can be other clients with similar data distributions~\cite{won2021fedmatch}, or the local, global, and the mixture of the local and global models by model splicing~\cite{che2021ssflnew}. However, the additional computational and communication overhead for the server to find and send the appropriate experts for each client can become exponentially costly with the increasing participating clients~\cite{wu2022fedcomm}. Moreover, these works treat all experts equally, taking an average of their knowledge for leveraging each client's unlabeled data. Due to this, we show that these methods' performance degrades significantly (see \Cref{sec:exprest}) when the number of labels decreases and the data heterogeneity gets higher. In our work, \textit{we use only the natural two experts that are available in FL, local and global, and show that this is enough to fully utilize the unlabeled data} at the clients when executed properly for \textit{both high and lower label scarcity and data heterogeneity cases}. Other related work proposes methods with restrictive assumptions such as the server having labeled data that is similar to the data distribution of the clients~\cite{diao2022ssfl} or several clients having fully labeled data~\cite{liang2022ssflnewfed}. In FL, the server does not have access to client data and labels are scarce, making these assumptions practically improbable. 

\renewcommand{\arraystretch}{0.9}
\begin{table*}[!t]\small
\setlength\tabcolsep{0.3pt}
\centering 
\begin{tabular}{lccccc}
\toprule
\multirow{2}{*}{Method} & Requires Server & Requires Fully & Requires Additionally & Requires Additional  & Robust to   \\
& Labeled Data & Labeled Clients & Computed Expert(s)  &  Comm.- of Params. & Class-Dist. Mismatch \\
\midrule
SemiFL~\cite{diao2022ssfl} & Yes  & No & No & No & No      \\
Rscfed~\cite{liang2022ssflnewfed} & No  & Yes & No & No & No      \\
FedTriNet~\cite{che2021ssflnew}  & No  & No & Yes & Yes & No  \\
FedMatch~\cite{won2021fedmatch}   & No  & No & Yes & Yes & No   \\
\textbf{FedLabel (ours)}  & \textbf{No} & \textbf{No} & \textbf{No} & \textbf{No} & \textbf{Yes}\\
\bottomrule
\end{tabular}
\caption{ Comparison of related work with \fedl.}
\label{table:comp}
\end{table*}

As summarized in \Cref{table:comp}, previous work in SSFL: i) assumes restrictive settings such as the server or several clients having good-enough fully labeled data, ii) imposes additional computational/communication burden at the server to find and send more experts other than the naturally occurring local and global models, and iii) does not consider the poor generalization of limited data to the unlabeled data such as class distribution mismatch. Improving on these drawbacks, we propose our novel method \fedl, which:
\begin{itemize}[leftmargin=*]
 \item Is robust to both the limited generalizability of the labeled data such as class distribution mismatch and data heterogeneity by using just two experts, global and local, to leverage a large number of unlabeled data ($80$-$95\%$) with just a few labeled data ($5$-$20\%$). \vspace{-0.3em}
 \item Does not require the server having any labeled data or a few clients to have fully labeled data. It also does not require additional experts to be computed or communicated other than the local and global model used in standard FL algorithms~\cite{mcmahan2017communication,sahu2019federated}. \vspace{-0.3em}
    \item Leverages unlabeled data by adaptively choosing either the local or global model based on the confidence of the model's prediction for pseudo-labeling with our proposed global-local consistency regularization that minimizes the divergence between the models' outputs when their pseudo-labels are identical. \vspace{-0.3em}
    \item Achieves $8$-$24\%$ test accuracy improvement compared to the other SSFL baselines, and even achieves a higher test accuracy than fully supervised scenario ($100\%$ labeled data) with only using $5$-$20\%$ of labels for extensive experiments (3 tasks for cross-device and 2 tasks for cross-silo).
\end{itemize} 
Before going into the details of our proposed method, we review the related literature on SSL and FL in the next section.

\section{Related Work} \vspace{-0.5em} \label{sec:related}
\noindent\textbf{Semi-Supervised Learning for General ML.} Labels are required to train well-performing models for classification tasks, but realistically labels are often scarce and expensive to obtain~\cite{lu2022unsupfl,lin2021flpos,cul2005labeling,yuji2021labeling,josep2017labeling}. To tackle such label scarcity for general ML problems, there has been a wide range of works including methods such as consistency regularization~\cite{xie2020uda,jeong2019consisssfl,verma2019intssfl}, pseudo-labeling~\cite{sohn2020fixmatch,zhang2021flexmatch,xu2021dash}, virtual adversarial training~\cite{miya2019vat}, and per-sample weighting of unlabeled data~\cite{ren2020notallssfl}. As one of the most popular methods, consistency regularization leverages unlabeled data by minimizing a model's prediction difference between the original image and perturbed versions of the image~\cite{xie2020uda,jeong2019consisssfl,verma2019intssfl}. Another commonly used method is pseudo-labeling, a simple approach to apply thresholding to the max probability of the prediction and provide a pseudo label for a data sample~\cite{sohn2020fixmatch}. Pseudo-labeling has also been popularly used in variations such as dynamic thresholding based on the sample's relatedness to the labeled data~\cite{xu2021dash}, or class-based thresholding~\cite{zhang2021flexmatch}. However, the aforementioned work does not jointly consider the poor generalizability that the limited labeled data can have to the unlabeled data and the data heterogeneity across clients. We show in \Cref{sec:exprest} that naively combining these SSL methods to FL significantly fails, while \fedl~is robust to both limited labeled data and data heterogeneity. \\ \vspace{-1em}

\noindent\textbf{Semi-Supervised Learning for FL.} Recently, several SSFL methods have been proposed such as utilizing inter-client consistency across clients~\cite{won2021fedmatch}, leveraging the labeled data at the server or several clients for communication-efficient or personalized SSL for FL~\cite{diao2022ssfl,he2021ssfl,liang2022ssflnewfed}, or using node classification on graphs for handling data with new label domains~\cite{wang2020graphfl}. FedMatch~\cite{won2021fedmatch} and FedTriNet~\cite{che2021ssflnew} are the closest to our work. In FedMatch, the server finds clients with similar data distributions for each client and sends the predictions from similar data clients for inter-client consistency regularization. In FedTriNet, each client handles three models (local, global, and the combination of the two through model splicing) to leverage the unlabeled data through a simple average of predictions across the three experts, and also requires the clients to send back their loss values to the server. Both approaches not only impose additional communication and computational costs for the clients and server but also their performance highly depends on finding the right parameter such as the number of helper clients or how to splice the models. Moreover, both methods do not consider weighing the experts differently to further combat limited data generalizability and data heterogeneity. We show in \Cref{sec:exprest} that both FedMatch and FedTrinet indeed fail further for a lower percentage of labeled data and higher data heterogeneity, while \fedl~in fact performs even better in these cases by simply leveraging only the local and global models. 


\vspace{-0.2em}
\section{Problem Formulation} \vspace{-0.5em}
We consider a FL setup (cross-silo and cross-device) where $M$ clients are connected to a central server to train a well-performing global model for a $N$-class classification task. Each client's local data, denoted as $\mathcal{D}_k~,k\in[M]$, is consisted of the labeled set $(\xb,y)\in\mathcal{D}_{L,k}$ and unlabeled set $\xi\in\mathcal{D}_{U,k}$, i.e., $\mathcal{D}_k=\mathcal{D}_{L,k}\bigcup\mathcal{D}_{U,k}$ where $\xb,~\xi\in\mathbb{R}^d$ is the input and $y\in[1,N]$ is the label. \\ \vspace{-0.7em}

\noindent\textbf{Conventional FL Algorithms' Assumption.} Previous work in FL~\cite{karimireddy2019scaffold,reddi2020adaptive} assumes an ideal scenario where each client $k$ has its unlabeled data $\mathcal{D}_{U,k}$ labeled which we call as the \textit{hypothetically labeled} unlabeled dataset denoted as $\mathcal{D}_{\overline{U},k}$. 
With the `hypothetical' fully labeled dataset of each client $k$ denoted as $\overline{\mathcal{D}}_{k}=\mathcal{D}_{L,k}\bigcup\mathcal{D}_{\overline{U},k}$, conventional FL algorithms assumes fully labeled data where the server aims to find the model parameter $\wb\in\mathbb{R}^q$ that minimizes: \vspace{-0.8em}
\begin{align} 
\gf(\wb)\hspace{-0.1em} & =\hspace{-0.1em} \sum_{\ic=1}^\ec \rdat F_k(\wb),~F_k(\wb)\defeq\frac{1}{|\overline{\mathcal{D}}_{k}|}\hspace{-0.1em}\sum_{\xi \in \overline{\mathcal{D}}_{k}} f(\wb, \xi) \nonumber
\end{align}
where $p_k$ is the aggregating weight and $f(\wb, \xi)$ is the loss function for sample $\xi\defeq(\xb,y)$ and parameter vector $\wb$. \\ \vspace{-1em}

\noindent\textbf{Realistic FL with Limited Labeled Data.} In practice, much of the available local data may not have ground-truth labels. In fact, the number of unlabeled data can be much larger than the labeled data, i.e., $|\mathcal{D}_{L,k}|\ll|\mathcal{D}_{U,k}|$. Then the server can only use the labeled data and effectively minimize\vspace{-0.8em}
\begin{align}\vspace*{-1.5em}
F_{\mathcal{L}}(\wb)\hspace{-0.2em}= \hspace{-0.2em}\sum_{\ic=1}^\ec \rdat \fsup(\wb),~\fsup(\wb)\hspace{-0.2em}\defeq\hspace{-0.2em}\frac{1}{|\mathcal{D}_{L,k}|}\hspace{-0.2em}\sum_{\xi \in \mathcal{D}_{L,k}} \hspace{-0.2em}f(\wb, \xi) \nonumber
\end{align}
where ${\wb^*_{\mathcal{L}}}=\argmin_{\wb}{F_{\mathcal{L}}}(\wb)$ becomes more different to the solution of the ideal objective $\wb^*=\arg\min_{\wb}{F(\wb)}$ as the distribution of $\bigcup_{k=1}^\ec\mathcal{D}_{L,k}$ differs more from the hypothetical ideal dataset $\bigcup_{k=1}^\ec\overline{\mathcal{D}}_{k}$. Our goal is to find an algorithm that can find the model parameter $\wb^*$ by using only the labels from $\bigcup_{k=1}^\ec\mathcal{D}_{L,k}$ and the unlabeled data $\bigcup_{k=1}^\ec\mathcal{D}_{U,k}$. \\ \vspace{-0.5em}

With clients having only a few labeled data and a larger number of unlabeled data, the labeled data can have limited generalization properties to the large unlabeled data due to factors such as class-distribution mismatch (the discrepancy between the distribution $\mathcal{D}_{L,k}$ and $\mathcal{D}_{\overline{U},k}$) or the mere limited number of its samples. Moreover, data heterogeneity (the discrepancy across the distributions of the clients' local data $\overline{\mathcal{D}}_k,~k\in[M]$) exacerbates the difficulty of SSFL. We show in our work that our proposed \fedl~enables clients to leverage their unlabeled data to the full extent just by selectively using the local and global model despite the difficulties caused by data heterogeneity and limited generalizability of the labeled data as explained in detail in the next section.


\section{\fedl: Choose Local or Global}
\label{sec:FedLabel}
\begin{figure*}[!h]
\centering
\includegraphics[width=1\textwidth]{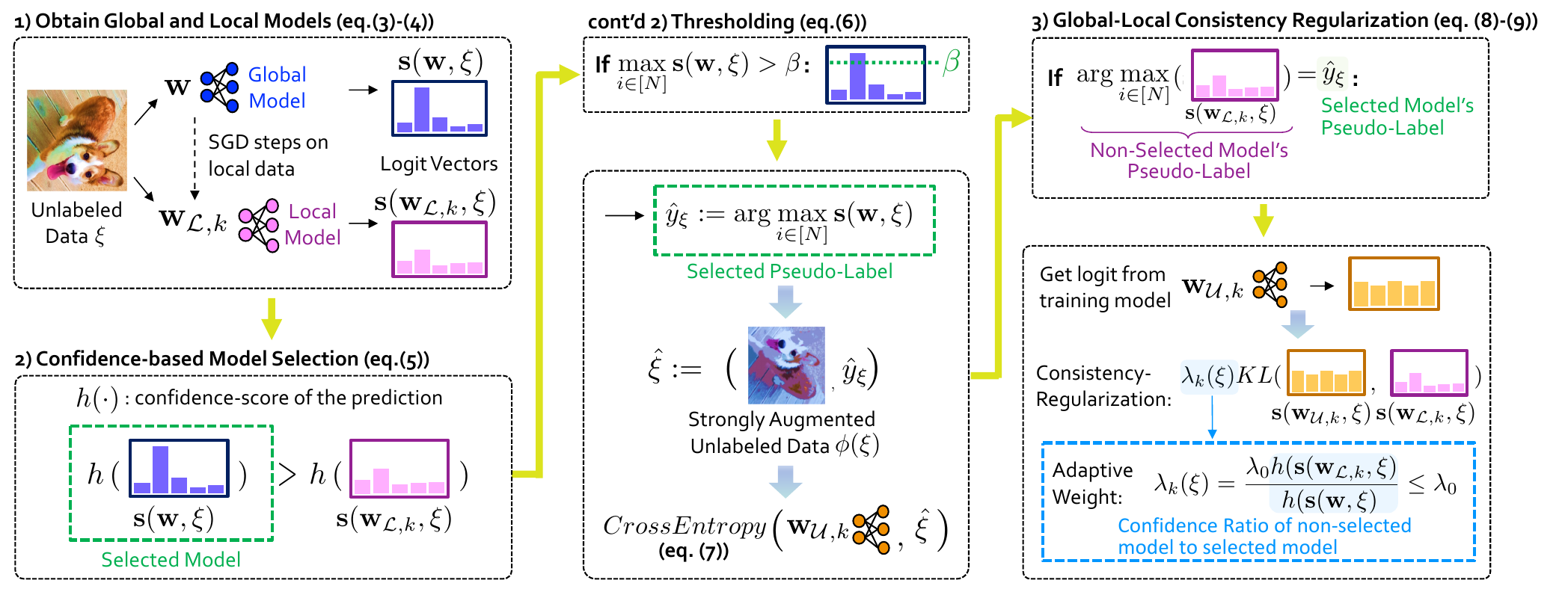}\caption{ Overview of \fedl~leveraging unlabeled data for each client $k\in[M]$. \fedl~consists of 3 main steps to leverage unlabeled data: 1) Obtaining global and local models (eq. \Cref{eqn:local_model_update}-\Cref{eq:suplocalupdate}), 2) Confidence-based selection with thresholding for obtaining the pseudo-label for the cross-entropy loss (eq. \Cref{eq:pseul0}-\Cref{eq:pseul}), and 3) the global-local consistency regularizing term to assimilate knowledge from both global and local models when applicable (eq. \Cref{eq:loc-glob1}-\Cref{eq:loc-glob2}). Details of the end-to-end algorithm of \fedl~is in \Cref{algo1}. 
}
\label{fig:overview}
\end{figure*}

In this section, we first introduce the novel semi-supervised loss function of \fedl~over the clients' unlabeled data and then present the end-to-end algorithm of \fedl~for implementation when used in realistic FL frameworks. An overview of how \fedl~leverages unlabeled data for each client $k\in[M]$ is in \Cref{fig:overview}.

\subsection{Semi-supervised Loss of \fedl.}  \vspace{-0.2em}

In the most commonly used vanilla FL~\cite{mcmahan2017communication}, there are two natural sources the clients can learn from: \textit{the global model} which is trained across different clients, and \textit{the local model} which is trained further with local SGD with their own local data. Whether the local or global model, or even both, can be effective for labeling the unlabeled data depends on which is more knowledgeable on the data based on what each has learned. For instance, if a client has very few labeled data, leading to class distribution mismatch or limited generalization capability to the unlabeled data, it can perform badly at correctly matching the labels for the unlabeled data. Instead, the global model can be more effective in giving the correct labels for the unlabeled data since it has seen more data from different clients. On the other hand, if the local labeled data sufficiently generalizes well to the local unlabeled data, the local model is more likely to give correct pseudo labels. This is also observed in \Cref{tab:locgloabl} where smaller number of labels leads to the `Only Global' performing better, but for larger number of labels `Only Local' performs better. There can also be cases where both the local and global models are not useful for leveraging the unlabeled data, and it is best not to use either model.

Based on this observation, we propose \fedl~that adaptively chooses either the local or global model based on the confidence score of each model's logits for pseudo-labeling. If both models' confidence does not exceed a certain threshold, we do not use that unlabeled data. Such binary choice of the model, however, can lead to losing relevant information from the other discarded model despite it having the same hard-label prediction as the chosen model. To assimilate both information from the local and global model for such scenarios, \fedl~adds global-local consistency regularization that minimizes the divergence between the local and global models' outputs. To the best of our knowledge, adaptive selection of the global and local model for pseudo-labeling and assimilating more knowledge, when needed, with global-local consistency regularization is a novel method that has not been previously proposed.\\ \vspace{-0.5em}

\noindent\textbf{1) Obtaining Global and Local Models (same as Standard FedAvg~\cite{mcmahan2017communication}).} With superscript $(t,r)$ denoting the communication round $t$ and local iteration $r$, for each $t$ the server selects a set of clients $\mathcal{C}^{(t,0)}$ uniformly at random and sends the global model $\wb^{(t,0)}$ to clients in $\mathcal{C}^{(t,0)}$. The clients in $\mathcal{C}^{(t,0)}$ initialize their local model for supervised training as $\wb_{\mathcal{L},\ic}^{(t,0)}=\wb^{(t,0)}$ to perform $\tau$ local iterations with learning rate $\lr$ to obtain their respective \textit{supervised} local models as:
\begin{align}
&\text{(Perform Local SGD on \textit{Labeled} Data)} \nonumber \vspace*{-1em} \\ &~~~~~~~~\wb_{\mathcal{L},\ic}^{(t,\tau)} =\wb_{\mathcal{L},\ic}^{(t,0)}-\lr\sum_{l=0}^{\tau-1}\nabla\fsup(\wb_{\mathcal{L},\ic}^{(t,l)},\xi_{\mathcal{L},\ic}^{(t,l)})
\vspace*{-1em}
\label{eqn:local_model_update}\\
&\text{(Compute the \textit{Supervised} Local Update)      }~~~~~~ \nonumber\\&~~~~~~~~\Delta \wb_{\mathcal{L},k}^{(t,0)} = \wb_{\mathcal{L},\ic}^{(t,\tau)}-\wb_{\mathcal{L},\ic}^{(t,0)} \label{eq:suplocalupdate}
\end{align}
where $\nabla\fsup(\wb_{\mathcal{L},\ic}^{(t,l)},\xi_{\mathcal{L},\ic}^{(t,l)})=\frac{1}{b}\sum_{\xi \in \xi_{\mathcal{L},\ic}^{(t,l)}}\nabla \lff(\wb_{\mathcal{L},\ic}^{(t,l)}, \xi)$ is the stochastic gradient computed with mini-batch $\xi_{\mathcal{L},\ic}^{(t,l)}$ of size $b$ randomly sampled from $\mathcal{D}_{L,k}$. Note that \fedl~does not alter the local update procedure used in standard FL algorithms~\cite{mcmahan2017communication,sahu2019federated}, and can be easily extended to using different methods to obtain the local model $\wb_{\mathcal{L},\ic}^{(t,\tau)}$.\\ \vspace{-0.5em}

\noindent\textbf{2) Confidence-based Selection with Thresholding.} After performing local SGD on the labeled data as in \Cref{eqn:local_model_update}, clients have the global model $\wb^{(t,0)}$ and the local model trained with labeled data $\wb_{\mathcal{L},\ic}^{(t,\tau)}$. Each client $k\in\mathcal{C}^{(t,0)}$ gets the logits from each the global and local model from the unlabeled data $\xi\in\dunl$ denoted as $\sbl(\wb^{(t,0)},~\xi)$ and $\sbl(\wb_{\mathcal{L},\ic}^{(t,\tau)},~\xi)$ respectively where $\mathbf{s}(\cdot,\cdot):\mathbb{R}^{q}\times\mathbb{R}^{d}\rightarrow\mathbb{R}^{N\times1}$. We then use function $h(\cdot):\mathbb{R}^{N\times1}\rightarrow \mathbb{R}$ that calculates the confidence-score (variance) of the logits and select the logit with the higher confidence-score to pseudo-label the unlabeled data. While we use variance to calculate confidence as in previous literature~\cite{yjc2022fedet}, \fedl~is not restricted to this metric. We include an ablation study on what to use as the confidence score of the logits in \Cref{app:exprest}. Formally, we have that
\begin{flalign}
&\begin{aligned}
     &\text{Binary Selection of Logit:}~\sbl^*(\xi)=\argmax_{\sbl\in\stl}h(\sbl),\\
     &~~~~~~~~~~\stl\defeq\{\sbl(\wb^{(t,0)},~\xi),\sbl(\wb_{\mathcal{L},\ic}^{(t,\tau)},~\xi)\}\end{aligned} \label{eq:pseul0}\\ 
&\begin{aligned}
     &\text{Pseudo Label from Thresholding:}\\
     &~~~~~~~~~~\psela_\xi(\beta)=\argmax_{i\in[N]}\sbl^*(\xi)\indi(\max\sbl^*(\xi)>\beta) \label{eq:pseul}
\end{aligned}
\end{flalign}
We discard the instances of $\hat{y}_\xi(\beta)=0$ which indicates that the selected logit did not pass the thresholding function in \Cref{eq:pseul}. Given that $\hat{y}_\xi(\beta)\neq0$, we have the pseudo-label $\psela_\xi(\beta)$ obtained via \Cref{eq:pseul} for the unlabeld data $\xi\in\mathcal{D}_{U,k}$. Then, with the separately set local model to be trained over the unlabeled data defined as $\mathbf{w}_{\mathcal{U},k}$, we have the cross-entropy loss for \fedl~as
\begin{align}
CrossEntropy(\mathbf{w}_{\mathcal{U},k},~\hat{\xi}),~\hat{\xi}\defeq (\psi(\xi),\psela_\xi(\beta)) \label{eq:ce}
\end{align}
where $\psi(\cdot)$ is a strong-augumentation~(RandAugment~\cite{cubuk2019randaug}) of the data sample $\xi$ with its hyperparameters set to $(1, 10)$. \\  \vspace{-0.5em}

\noindent\textbf{3) Global-Local Consistency Regularization.} The cross-entropy loss of \fedl~in \Cref{eq:ce} uses the pseudo-label selected via a binary selection between the local and global model which essentially discards the logits of the model that has not been selected. However, there can be cases where the discarded model's soft logits which we denote as $\sbl^{-*}(\xi)$ also point to the same label $\psela_{\xi}$, i.e., $\argmax_{i\in[N]}\sbl^{-*}(\xi)=\psela_\xi(\beta)$ as the selected model. In such cases with only the cross-entropy term in \Cref{eq:ce}, \fedl~can lose the useful information contained in the discarded model. To cover such scenarios, \fedl~utilizes the knowledge of the discarded model when it predicts the same label as the selected model by our proposed global-local consistency regularizing term as follows:
\begin{align}
&\lambda_k(\xi) KL(\sbl(\mathbf{w}_{\mathcal{U},k},\xi),\sbl^{-*}(\xi))\indi(\argmax_{i\in[N]}\sbl^{-*}(\xi)=\psela_\xi(\beta)), \label{eq:loc-glob1}\\
&~\text{where}~\lambda_k(\xi)\defeq\lambda_0 h(\sbl^{-*}(\xi))/h(\sbl^{*}(\xi)) \label{eq:loc-glob2}
\end{align}
Note that the global-local consistency regularizing term (KL-divergence) is weighed by $\lambda_k(\xi)$ which is the confidence-score ratio by the discarded model to the selected model, i.e., $\lambda_0 h(\sbl^{-*}(\xi))/h(\sbl^{*}(\xi))\leq \lambda_0$, so that the less confident the discarded model is compared to the selected model, the lower the regularization weight. The maximum value of $\lambda_k(\xi),~\xi\in\mathcal{D}_{\mathcal{U},k},~\forall k\in[M]$ is $\lambda_0$ which is achieved when the confidence score of the discarded model and the selected model is identical. We show in \Cref{sec:exprest} that this consistency regularization term indeed helps \fedl~improve its performance from at least $5\%$ to at most $9\%$. \\ \vspace{-0.5em}


\noindent\textbf{\fedl's Final Semi-supervised Loss.} Combining the cross-entropy loss from the confidence-based selection in \Cref{eq:ce} and the global-local consistency regularization loss in \Cref{eq:loc-glob1}, we have the final semi-supervised loss of \fedl~denoted as $\fun(\wb)$ for each client $k\in[M]$ as below:
\begin{flalign}
\begin{aligned}
&\fun(\mathbf{w}_{\mathcal{U},k})=\frac{1}{|\mathcal{D}_{\mathcal{U},k}|}\hspace{-0.3em}\sum_{\xi\in\mathcal{D}_{\mathcal{U},k}}\hspace{-0.5em}(CrossEntropy(\mathbf{w}_{\mathcal{U},k},~\hat{\xi})\\
~~~~&+\hspace{-0.2em}\lambda_k(\xi) KL(\sbl(\mathbf{w}_{\mathcal{U},k},\xi),\sbl^{-*}(\xi))\indi(\argmax_{i\in[N]}\sbl^{-*}(\xi)\hspace{-0.2em}=\hspace{-0.2em}\psela_\xi(\beta))) \label{eq:unsuploss}
\end{aligned}
\end{flalign}
With the final semi-supervised loss of \fedl~proposed in \Cref{eq:unsuploss}, now we are ready to elaborate on how \fedl~is implemented in the subsequent subsection.

\setlength{\textfloatsep}{15pt plus 1.0pt minus 2.0pt}
\begin{algorithm} [!t]
\caption{ \fedl~Framework}\label{algo1}
\renewcommand{\algorithmicloop}{\textbf{Global server do:}}
\begin{algorithmic}[1]
\STATE {\bfseries Initialize} $\wb^{(0,0)}$,~{\bfseries Output:} Global model $\wb^{(T,0)}$,
\STATE {\bfseries For ${t=0,...,T-1}$ communication rounds do}:
\STATE \hspace*{1em} {\bfseries Global server: }Select $m$ clients for $\mathcal{C}^{(t,0)}$ uniformly at random and send $\wb^{(t,0)}$ to clients in $\mathcal{C}^{(t,0)}$
\STATE \hspace*{1em} {\bfseries Clients $\ic\in\mathcal{C}^{(t,0)}$ in parallel do:}
\STATE \hspace*{2em} {Set $\wb_{\mathcal{L},\ic}^{(t,0)}=\wb^{(t,0)},~\wb_{\mathcal{U},\ic}^{(t,0)}=\wb^{(t,0)}$}\\
\STATE {\hspace*{2em} Get $\wb_{\mathcal{L},\ic}^{(t,\tau)} =\Delta\wb_{\mathcal{L},k}^{(t,0)}+\wb_{\mathcal{L},\ic}^{(t,0)}$ (See \Cref{eq:suplocalupdate})}\\
\STATE {\hspace*{2em} Get $\Delta\wb_{\mathcal{U},k}^{(t,0)}$ (See \Cref{eq:unsuplocalupdate}}\\
\STATE {\hspace*{2em} Send $\Delta \wb_\ic^{(t,0)}=\Delta\wb_{\mathcal{L},k}^{(t,0)}+\Delta\wb_{\mathcal{U},k}^{(t,0)}$ and aggregation weight $r_\ic^{(t,0)}$ to the server}
\STATE {\hspace*{1em} {\bfseries Global server:}} Update $\wb^{(t+1,0)}=\wb^{(t,0)}+\sum_{k\in\mathcal{C}^{(t,0)}}\frac{r_k^{(t,0)}}{\sum_{k'\in\mathcal{C}^{(t,0)}}r_{k'}^{(t,0)}}\Delta\wb_{k}^{(t,0)}$\\ 
\end{algorithmic} 
\end{algorithm}

\subsection{\fedl~Implementation} 
 Recall that at each communication round $t$, the selected set of clients $\mathcal{C}^{(t,0)}$ initialize their local model to the global model, i.e., $\wb_{\mathcal{L},\ic}^{(t,0)}=\wb^{(t,0)}$ and perform $\tau$ local iterations to obtain the supervised local update $\Delta \wb_{\mathcal{L},k}^{(t,0)}$ (see \Cref{eq:suplocalupdate}). Using the same initialization $\wb_{\mathcal{U},\ic}^{(t,0)}\defeq\wb^{(t,0)}$, the clients now obtain the semi-supervised local update by minimizing the semi-supervised loss $\fun(\mathbf{w}_{\mathcal{U},k})$ (see \Cref{eq:unsuploss}) via mini-batch SGD for $\tau'$ iterations on the unlabeled data $\mathcal{D}_{\mathcal{U},k},k\in\mathcal{C}^{(t,0)}$. Concretely, we have the semi-supervised local update as:
 \vspace*{-0.5em}
\begin{align}
&\text{(Perform Local SGD on \textit{Unlabeled} Data)  } \nonumber \\  \vspace{-1em}&~~~~~~~~\wb_{\mathcal{U},\ic}^{(t,\tau)} =\wb_{\mathcal{U},\ic}^{(t,0)}-\lr\sum_{l=0}^{\tau'-1}\nabla\fun(\wb_{\mathcal{U},\ic}^{(t,l)},\xi_{\mathcal{U},\ic}^{(t,l)})
\\
&\text{(Compute the \textit{Semi-supervised} Local Update) } \nonumber \\&~~~~~~~~\Delta \wb_{\mathcal{U},k}^{(t,0)} = \wb_{{\mathcal{U},\ic}}^{(t,\tau)}-\wb_{\mathcal{U},\ic}^{(t,0)} \label{eq:unsuplocalupdate}
\end{align}
where $\nabla\fun(\wb_{\mathcal{U},\ic}^{(t,l)},\xi_{\mathcal{U},\ic}^{(t,l)})=\frac{1}{b}\sum_{\xi \in \xi_{\mathcal{U},\ic}^{(t,l)}}\nabla \lff(\wb_{\mathcal{U},\ic}^{(t,l)}, \xi)$ is the stochastic gradient computed using a mini-batch $\xi_{\mathcal{U},\ic}^{(t,l)}$ of size $b$ that is randomly sampled from client $\ic$'s local unlabeled dataset $\mathcal{D}_{U,k}$. Now each client $k\in\mathcal{C}^{(t,0)}$ sends back its update $\Delta\wb_{k}^{(t,0)}=\Delta\wb_{\mathcal{L},k}^{(t,0)}+\Delta\wb_{\mathcal{U},k}^{(t,0)}$ back to the server along with the total number of data samples $r_k^{(t,0)}$ used for obtaining the local updates $\Delta\wb_{\mathcal{L},k}^{(t,0)}$ and $\Delta\wb_{\mathcal{U},k}^{(t,0)}$. Note that $r_k^{(t,0)}$ is dependent on $t$ because the number of unlabeled data samples that are used for training varies for each communication round depending on how many samples pass the confidence-based threshold in \Cref{eq:pseul}. Then finally the server updates its global model as $\wb^{(t+1,0)}=\wb^{(t,0)}+\sum_{k\in\mathcal{C}^{(t,0)}}\frac{r_k^{(t,0)}}{\sum_{k'\in\mathcal{C}^{(t,0)}}r_{k'}^{(t,0)}}\Delta\wb_{k}^{(t,0)}$. The details of \fedl's implementation is in \Cref{algo1}.

\section{Experiments}\vspace{-0.2em}
\label{sec:exp}




\subsection{Experimental Setup} \vspace{-0.5em}
We perform experiments on a wide variety of experiments on \textbf{both cross-device and cross-silo settings, partial and full client participation, low and high data heterogeneity, and low and large label scarcity}. We perform 3 tasks for cross-device: Resnet18 with EMNIST (62 labels)~\cite{cohen2017emnist}, Resnet34 with CIFAR10 (10 labels), and Resnet50 with CIFAR100 (100 labels)~\cite{krizhevsky2009cifar} and 2 tasks for cross-silo: OrganAMNIST (11 labels) and BloodMNIST (8 labels)~\cite{medmnistv2} both with Resnet18. For cross-device, we select $10\%$ of clients for training for each comm. round. For the OrganAMNIST and BloodMNIST we select $30\%$ and $100\%$ of clients. \\ \vspace{-1em}

\noindent \textbf{Data Partitioning Across Clients.} For cross-device, the data is partitioned across 100 clients using the Dirichlet distribution $\text{Dir}(\alpha=0.1)$ \cite{hsu2019noniid}, unless mentioned otherwise. The parameter $\alpha$ determines the degree of data heterogeneity (smaller $\alpha$ indicates larger data heterogeneity). For cross-silo, data is partitioned across 10 and 5 clients in total respectively with $\alpha=0.1$, unless mentioned otherwise. Note that we emulate a realistic setting where the total number of clients is much smaller for cross-silo than for cross-device. \\ \vspace{-1em}

\noindent \textbf{Data Partitioning Within Clients.} The local training dataset of each client is partitioned into labeled and unlabeled data uniformly at random. Note that even without artificially biasing the dataset partitioning, we were still able to observe class-distribution mismatch for high label scarcity (see \Cref{fig:moti2}). For cross-device, we have $20\%:80\%$ and $50\%:50\%$ partitioning of the labeled : unlabeled data for each client's local training data. For cross-silo, we have $5\%:95\%$ and $20\%:80\%$ partitioning. Note that we have run more fine-grained data ratio experiments (10,20,50,65 for CIFAR10, and 5,20,35,50 for OrganAMNIST), {also shown in \Cref{fig:moti}}, but have selected a few intervals that have shown the most significant difference for presentation.  \\ \vspace{-1em}

\noindent \textbf{Baselines.} We compare \fedl~with 3 classes of baselines: i) Supervised FL baselines with Fully Labeled Data denoted as $100\%$ (\textbf{FedAvg (}$\mathbf{100\%}$\textbf{)}~\cite{mcmahan2017communication}, \textbf{FedProx (}$\mathbf{100\%}$\textbf{)}~\cite{sahu2019federated}), ii) Supervised FL baselines with Partially Labeled Data (\textbf{FedAvg, FedProx}), and iii) SSFL baselines~\cite{xie2020uda,sohn2020fixmatch} with Partially Labeled Data ((\textbf{FedAvg+UDA, FedAvg+FixMatch, FedProx+UDA, FedProx+FixMatch, FedTriNet~\cite{che2021ssflnew}, FedMatch~\cite{won2021fedmatch}}). The i) supervised FL baselines with $100\%$ of labeled data is the hypothetical \textit{upper bound} that we can achieve by using FL when clients have all their data labeled, and ii) the supervised FL baselines with partially labeled data is the \textit{lower bound} that FL actually achieves in a realistic setting with only a few labeled data. The iii) SSFL baselines are the state-of-the-art methods that tackle label scarcity in FL. We do not compare with SemiFL~\cite{diao2022ssfl} and Rscfed~\cite{che2021ssflnew} which impose assumptions such as the server having labeled data or several clients having fully labeled data (see \Cref{table:comp}).

\begin{table*}[!t]  \centering 
\setlength\tabcolsep{3.5pt} \small
\begin{tabular}{@{}c:l||ccc||cc@{}} \dtoprule 
\multicolumn{2}{c||}{\multirow{2}{*}{\vspace*{0.5em} \shortstack{High Client \\ Data Heterogeneity}}} & \multicolumn{3}{c||}{Cross-Device Setting} & \multicolumn{2}{c}{Cross-Silo Setting} \\ 
\cmidrule(lr){3-5} \cmidrule(lr){6-7}
\multicolumn{2}{c||}{}  & {EMNIST} & {CIFAR10} & {CIFAR100} & {OrganAMNIST} & {BloodMNIST} \\  \cmidrule(lr){1-2} \cmidrule(lr){3-5} \cmidrule(lr){6-7}
Supervised, & FedAvg ($100\%$) & $78.36~{\scriptstyle (\pm 0.73)}$ & $44.67~{\scriptstyle (\pm 1.09)}$ & $24.67~{\scriptstyle (\pm 0.55)}$ & $71.09~{\scriptstyle (\pm 1.00)}$  & $70.53~{\scriptstyle (\pm 1.13)}$  \\  
Fully Labeled   & FedProx ($100\%$) & $\mathbf{78.60}~{\scriptstyle (\pm 0.89)}$ & $\mathbf{44.72}~{\scriptstyle (\pm 1.63)}$ & $\mathbf{24.87}~{\scriptstyle (\pm 0.19)}$ & $\mathbf{73.34}~{\scriptstyle (\pm 0.86)}$ & $\mathbf{74.53}~{\scriptstyle (\pm 0.39)}$ \\  \hdashline
Supervised, &  FedAvg & $63.05~{\scriptstyle (\pm 1.52)}$ & $28.83~{\scriptstyle (\pm 2.94)}$ & $11.33~{\scriptstyle (\pm 0.22)}$ & $41.90~{\scriptstyle (\pm 2.18)}$ & $53.18~{\scriptstyle (\pm 0.11)}$ \\  
Partially Labeled & FedProx & $63.31~{\scriptstyle (\pm 0.16)}$ & $31.09~{\scriptstyle (\pm 1.15)}$ & $11.37~{\scriptstyle (\pm 0.15)}$ &  $48.28~{\scriptstyle (\pm 0.78)}$ & $56.53~{\scriptstyle (\pm 0.68)}$ \\  \hdashline

\multirow{ 6}{*}{\shortstack{Semi-Supervised, \\ Partially Labeled}} & FedAvg+UDA & $68.98~{\scriptstyle (\pm 1.06)}$ & $31.32~{\scriptstyle (\pm 2.84)}$ & $10.80~{\scriptstyle (\pm 0.57)}$ & $55.14~{\scriptstyle (\pm 0.92)}$ & $53.18~{\scriptstyle (\pm 0.68)}$  \\  
& FedAvg+FixMatch & $67.98~{\scriptstyle (\pm 1.59)}$ & $23.99~{\scriptstyle (\pm 2.73)}$ & $10.16~{\scriptstyle (\pm 0.20)}$ & $52.76~{\scriptstyle (\pm 2.95)}$ & $47.92~{\scriptstyle (\pm 1.55)}$  \\  
 & FedProx+UDA & $71.93~{\scriptstyle (\pm 1.37)}$ & $31.99~{\scriptstyle (\pm 2.26)}$ & $10.57~{\scriptstyle (\pm 0.13)}$ & $54.15~{\scriptstyle (\pm 2.53)}$  & $59.48~{\scriptstyle (\pm 1.17)}$ \\  
  & FedProx+FixMatch & $70.56~{\scriptstyle (\pm 0.96)}$ &  $22.38~{\scriptstyle (\pm 2.28)}$ & $10.02~{\scriptstyle (\pm 0.54)}$ & $55.49~{\scriptstyle (\pm 2.65)}$  & $50.88~{\scriptstyle (\pm 1.12)}$   \\ 
   & FedTriNet & $60.31~{\scriptstyle (\pm 0.25)}$  & $29.56~{\scriptstyle (\pm 1.53)}$ & $10.53~{\scriptstyle (\pm 1.83)}$ & $50.82~{\scriptstyle (\pm 1.21)}$  & $59.38~{\scriptstyle (\pm 1.05)}$  \\ 
  & FedMatch & $63.48~{\scriptstyle (\pm 0.57)}$ & $31.94~{\scriptstyle (\pm 1.78)}$ & $10.86~{\scriptstyle (\pm 1.33)}$ & $48.99~{\scriptstyle (\pm 1.38)}$ & $58.89~{\scriptstyle (\pm 2.24)}$   \\ 
& \fedl~(\textbf{ours}) & $\mathbf{79.33}~{\scriptstyle (\pm 1.97)}$ & $\mathbf{46.05}~{\scriptstyle (\pm 1.09)}$ & $\mathbf{18.42}~{\scriptstyle (\pm 1.46)}$ & $\mathbf{69.43}~{\scriptstyle (\pm 1.58)}$ & $\mathbf{71.46}~{\scriptstyle (\pm 1.89)}$  \\ 
 \dbottomrule  \end{tabular}\caption{ Test accuracy for high label scarcity on each client's local data ($20$\% of labeled data for cross-device and $5\%$ for cross-silo). \fedl~achieves a significantly higher test accuracy by approximately $8$-$16\%$,~$15$-$24\%$,~$8\%$,~$14$-$20\%$, and~$12$-$23\%$ for EMNIST, CIFAR10, CIFAR100, OrganAMNIST, and BloodMNIST respectively. For EMNIST, CIFAR10, and BloodMNIST, \fedl~achieves \textit{an even higher test accuracy} than the supervised fully labeled case ($100\%$ of the data labeled) by approximately $1\%, 2\%$, and $1\%$. \label{tab:testacc} } 
\end{table*}

\begin{figure*}[!t]
\centering
\begin{subfigure}{0.49\textwidth}
\centering
\includegraphics[width=1\textwidth]{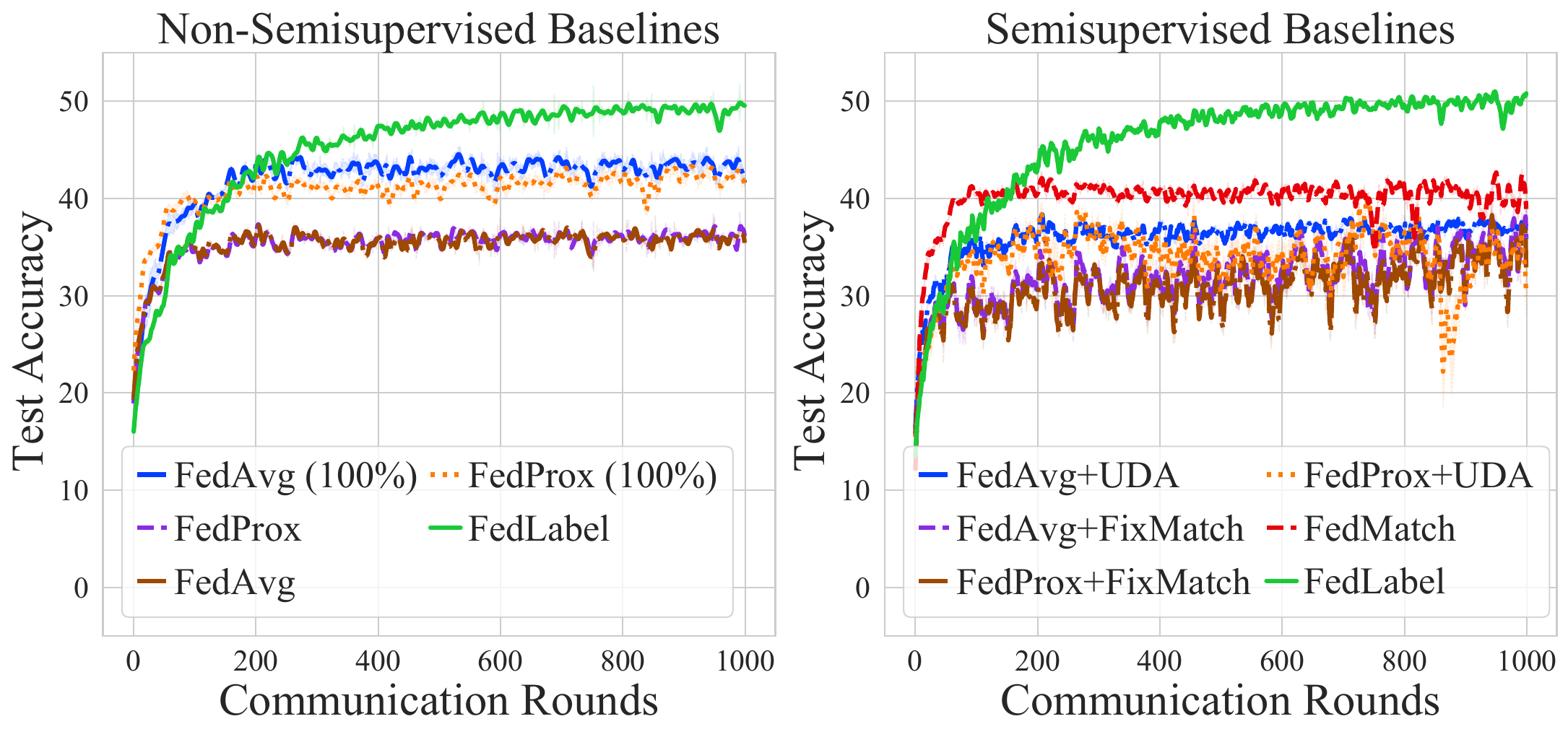} \vspace*{-1.7em} 
 \caption{ $50\%$ of Labeled Data for CIFAR10}
\end{subfigure} \hfill
\begin{subfigure}{0.49\textwidth}
\centering
\includegraphics[width=1\textwidth]{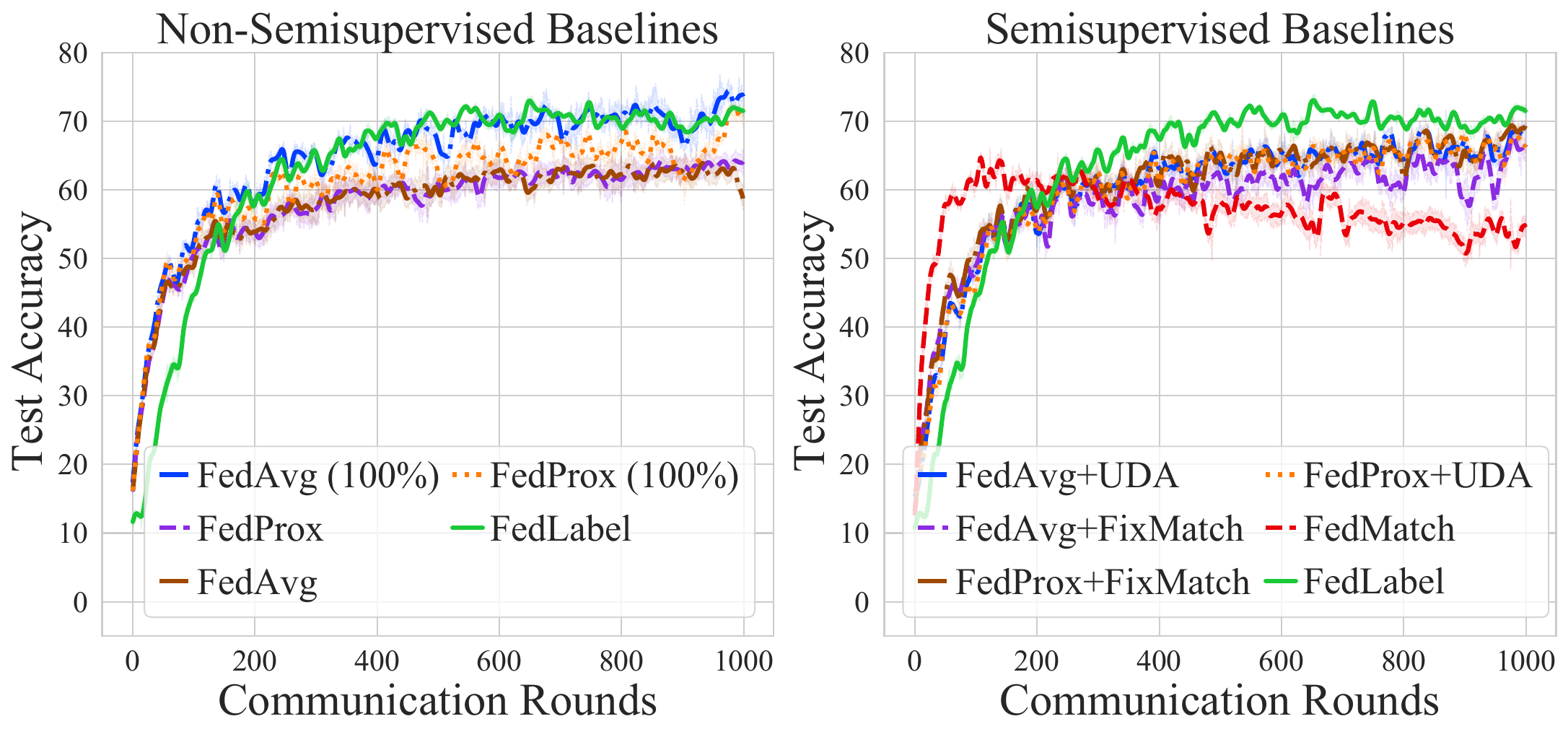}  \vspace*{-1.7em} 
 \caption{ $20\%$ of Labeled Data for OrganAMNIST}
\end{subfigure}
 \caption{ Test accuracy for lower label scarcity, $50\%$ and $20\%$ of labeled data, for each client's local training data for CIFAR10 and OrganAMNIST respectively. \fedl~outperforms the baselines by $8$-$15\%$ and $2$-$10\%$ for CIFAR10 and OrganAMNIST respectively. Comparing to results in \Cref{tab:testacc}, \fedl~performance gap with other baselines is higher when there is higher label scarcity.}
\label{fig:more_label}  
\end{figure*}

\subsection{Experimental Results} \label{sec:exprest}
We thoroughly evaluate \fedl~in the following aspects: achieves high accuracy for i) both low and high label scarcity ($5$-$20\%$ and $20$-$50\%$), and ii) both low and high data heterogeneity. We also perform ablation studies on \fedl~as follows: iii) effect of the global-local consistency regularizing term modulated by $\lambda_0$ in \Cref{eq:loc-glob2}, iv) effect of the number of local steps $\tau$ to obtain the local model in \Cref{eqn:local_model_update},  v) effect of the thresholding parameter $\beta$ in \Cref{eq:pseul}, and vi) effect of different confidence measures $h(\cdot)$ in \Cref{eq:pseul0}. We defer the results for v) and vi) to \Cref{app:exprest} due to space constraints.\\ \vspace{-1.9em}

\noindent \textbf{Effectiveness of \fedl~for High Label Scarcity.} We evaluate \fedl~with the top-1 test accuracies for $20\%$ and $5\%$ of labeled data on each client for cross-device and cross-silo respectively in \Cref{tab:testacc}. Compared to the other SSFL baselines, \fedl~achieves a higher test accuracy by $8$-$16\%$,~$15$-$24\%$,~$8\%$,~$14$-$20\%$, and~$12$-$23\%$ for EMNIST, CIFAR10, CIFAR100, OrganAMNIST, and BloodMNIST respectively. Surprisingly, for EMNIST, CIFAR10, and BloodMNIST, \textbf{\fedl~achieves an even higher test accuracy than the fully supervised case} ($100\%$ of labeled data) by approximately $1\%, 2\%$, and $1\%$. In other words, \fedl~performs even better than the ideal scenario when $100\%$ of the data is labeled, with only $5$-$20\%$ of labeled data. We reason that this is due to the small amount of noise introduced by \fedl~when leveraging the unlabeled data which can improve the generalization performance as shown in previous work~\cite{yu2019dis,gui2021noisy}. \fedl's also outperforms by $7$-$27\%$ than the supervised baselines with partially labeled data, while some of the SSFL baselines with partially labeled data perform even worse. Next, we evaluate \fedl~when there are more labels in the subsequent paragraph. \\ \vspace{-1em}

\noindent \textbf{Comparison with Cases for Lower Label Scarcity.} In \Cref{fig:more_label}, we show results for larger portions of labeled data, $50\%$ for CIFAR10 and $20\%$ for OrganAMNIST, than the results shown in \Cref{tab:testacc} which was for $20\%$ and $5\%$ of labeled data respectively. For the larger number of labeled data, \fedl~still outperforms the baselines by $8$-$15\%$ and $2$-$10\%$ for CIFAR10 and OrganAMNIST respectively. However, the performance gap is lower than the $15$-$24\%$ and $14$-$20\%$ improvement shown in \Cref{tab:testacc} for higher label scarcity. Hence, this shows that while \fedl~still outperforms other baselines for lower label scarcity, it outperforms the other baselines with a higher gap when clients have a smaller number of labels, i.e., high label scarcity.  \\ \vspace{-1em}

\begin{figure*}[!t]
\centering
\begin{subfigure}{0.243\textwidth}
\centering
\includegraphics[width=1\textwidth]{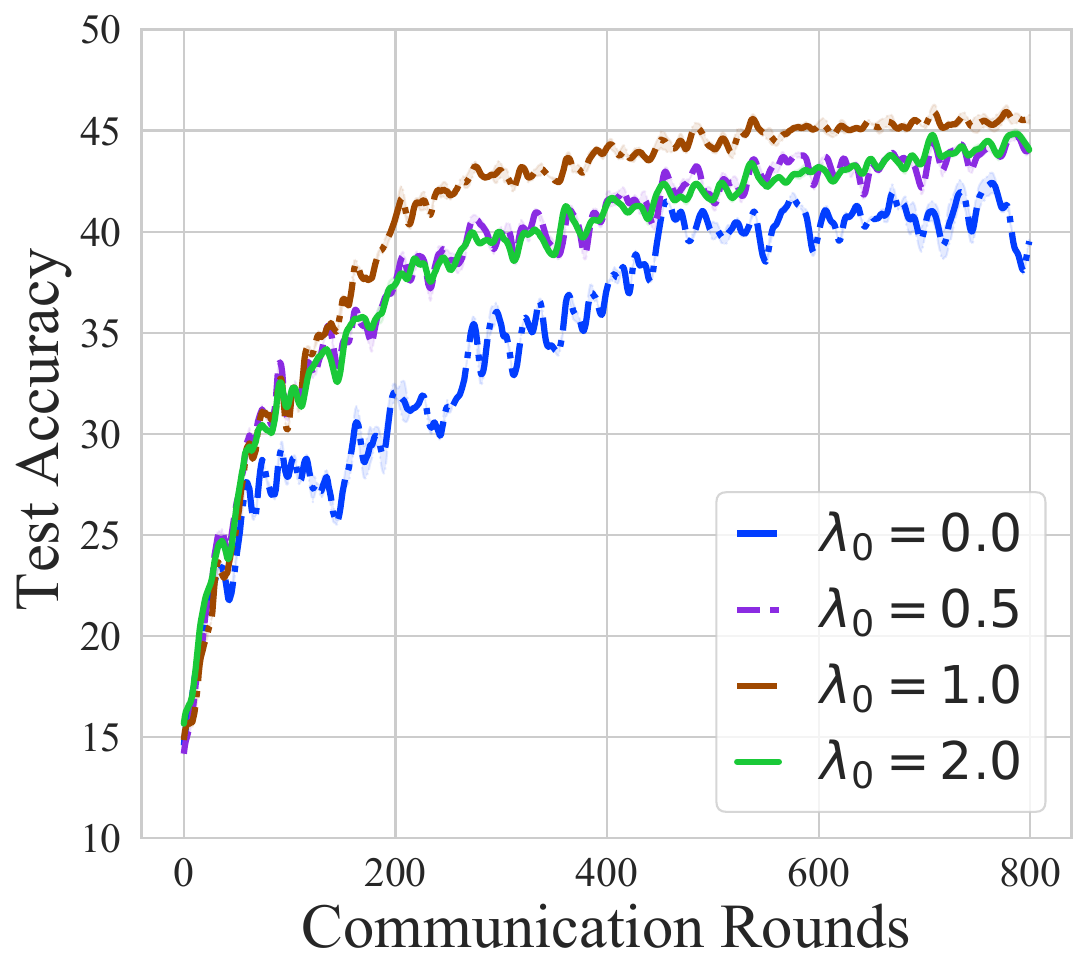}  \caption{ CIFAR10}
\end{subfigure} \hfill
\begin{subfigure}{0.243\textwidth}
\centering
\includegraphics[width=1\textwidth]{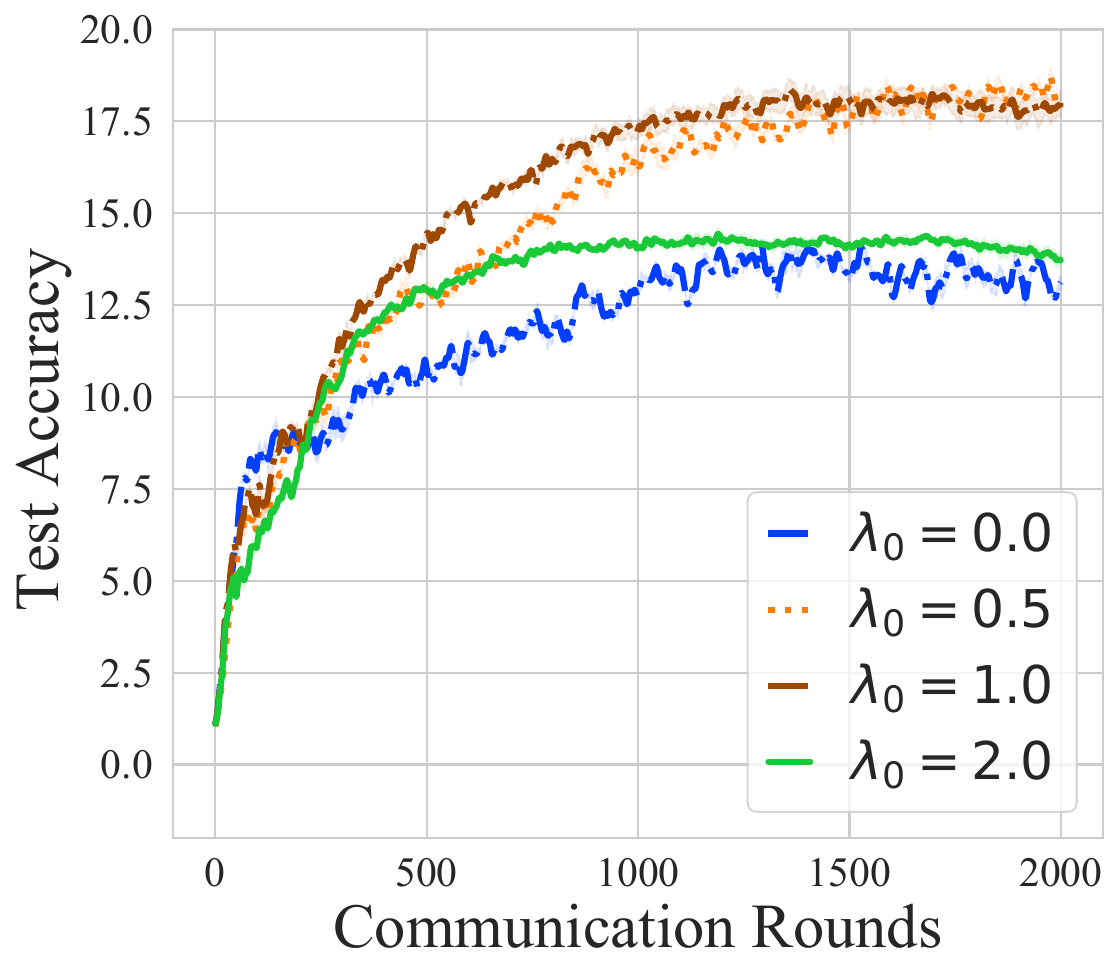}  \caption{ CIFAR100}
\end{subfigure} \hfill
\begin{subfigure}{0.243\textwidth}
\centering
\includegraphics[width=1\textwidth]{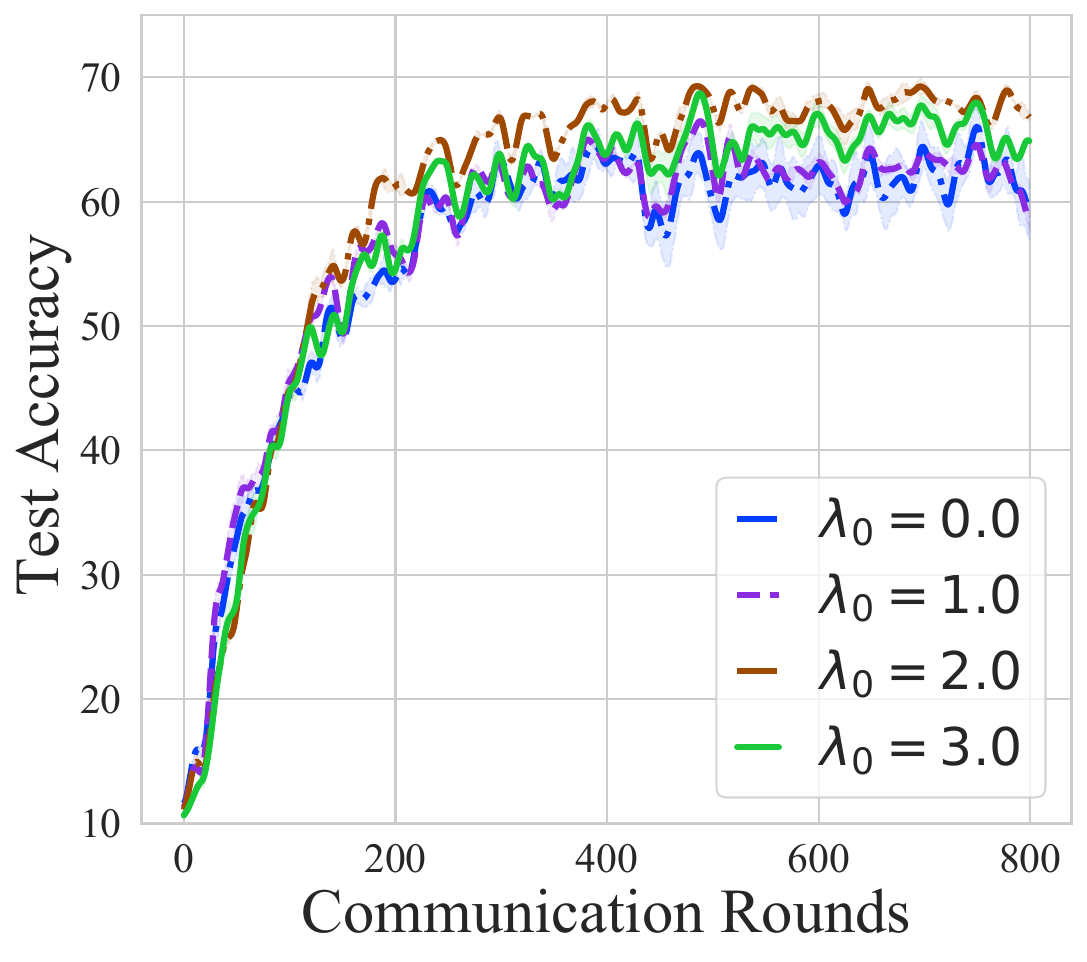} \caption{ OrganAMNIST}
\end{subfigure} \hfill
\begin{subfigure}{0.243\textwidth}
\centering
\includegraphics[width=1\textwidth]{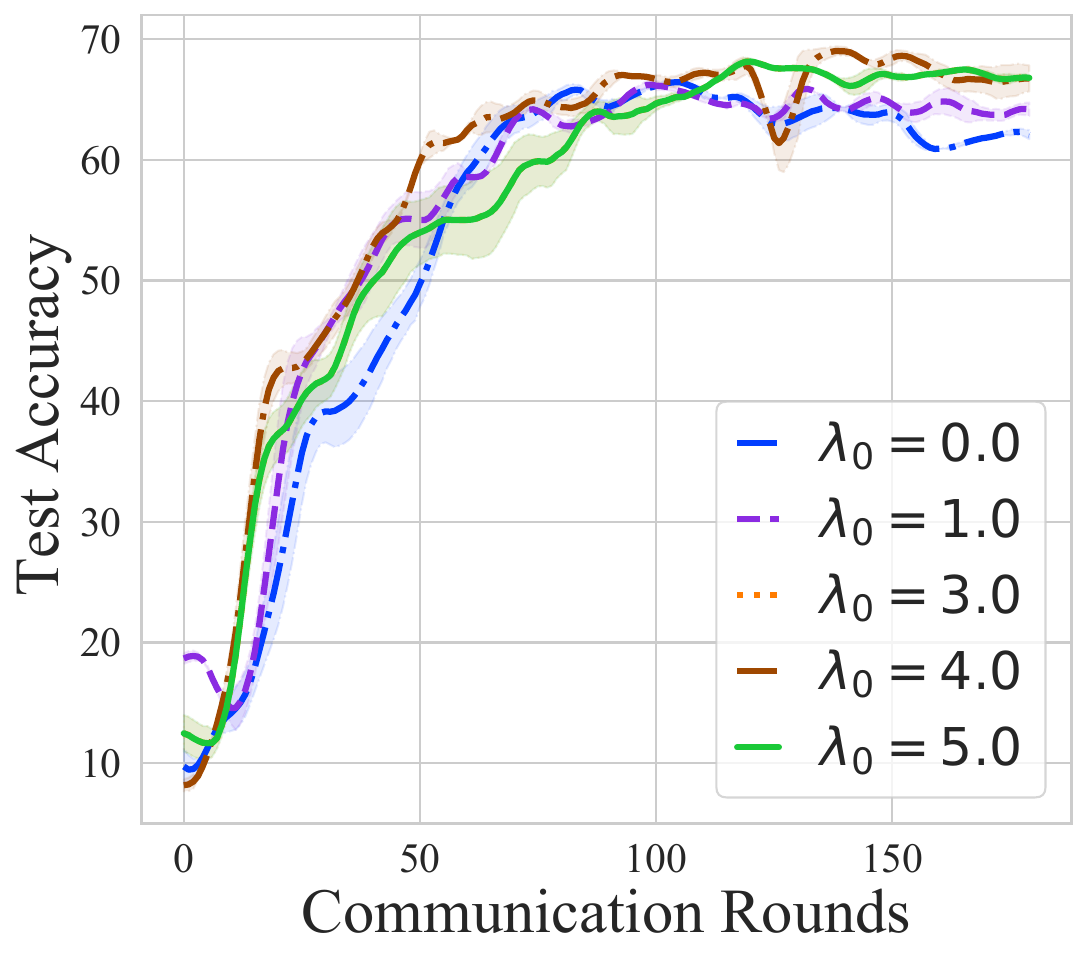}  \caption{ BloodMNIST}
\end{subfigure} \hfill
 \caption{ Ablation study on the effect of the global-local consistency regularizing term by modulating $\lambda_0$ in \Cref{eq:loc-glob2}. For all datasets, $\lambda_0=0$ gives the lowest test accuracy, showing that without the regularizing term, we lose useful information from the discarded model from the binary selection between the local and global model. For larger $\lambda_0>0$, the test accuracy improves by approximately $5$-$9\%$.} 
\label{fig:globlocabl}  
\end{figure*}
\comment{
\begin{figure*}[!t]
\centering
\begin{subfigure}{0.497\textwidth}
\centering
\includegraphics[width=1\textwidth]{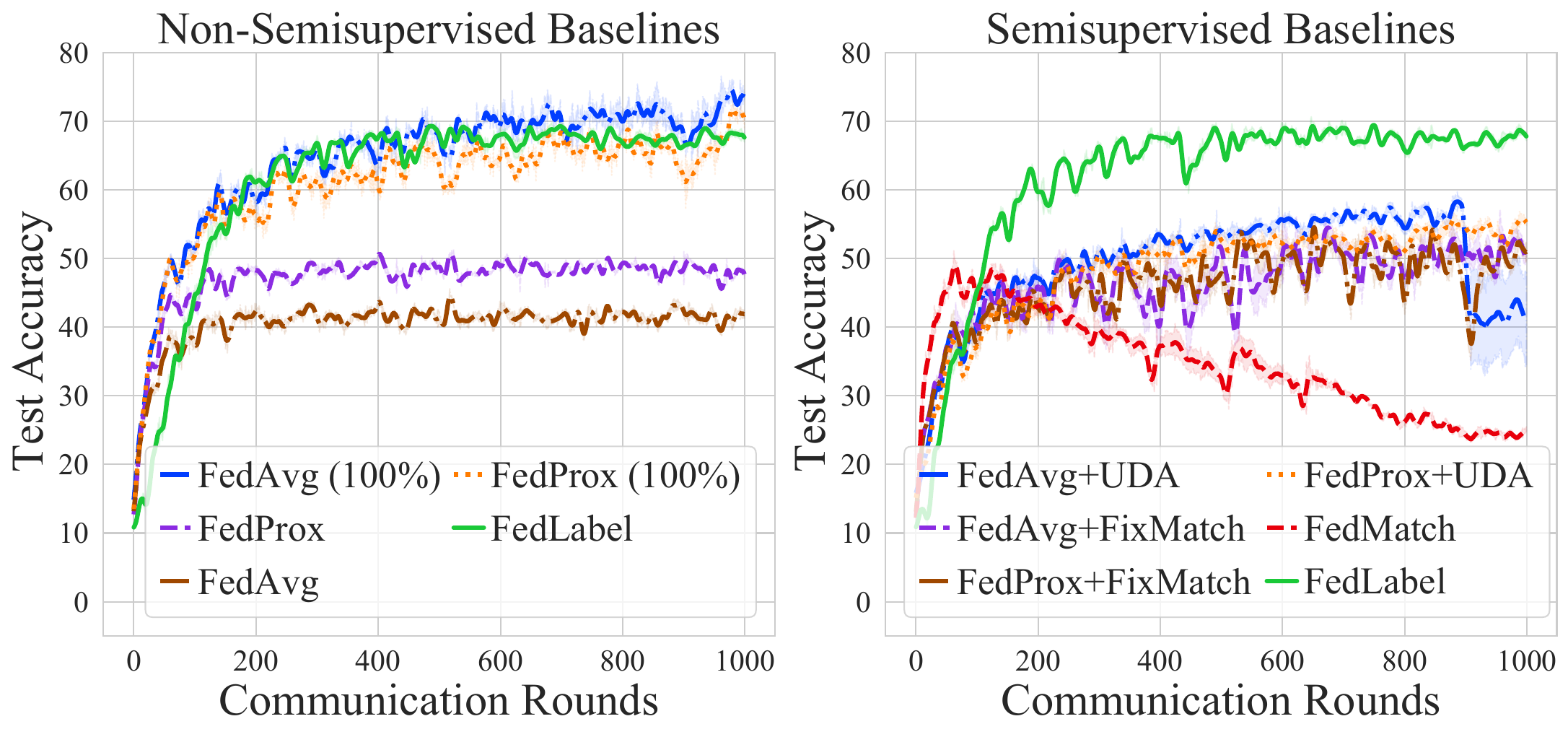} \caption{ $5\%$ of Labeled Data}
\end{subfigure} \hfill
\begin{subfigure}{0.497\textwidth}
\centering
\includegraphics[width=1\textwidth]{fig/ORGAN_L_2.pdf} \caption{ $20\%$ of Labeled Data}
\end{subfigure} 
 \caption{ Test accuracy for $5\%$ and $20\%$ of labeled data for each client's local training data for OrganAMNIST. \fedl~achieves nearly similar performance with the fully-supervised baselines ($100\%$ labeled data) with only $5\%,~20\%$ of labeled data respectively and~achieves $3$-$45\%$ higher test accuracy than the other SSL baselines.}
\label{fig:organ_l} 
\end{figure*}
}

\newfloatcommand{capbtabboxis}{table}[][0.47\textwidth]
\begin{figure}[!t]
\begin{floatrow}
\capbtabboxis{\renewcommand{\arraystretch}{1} 
\centering 
\setlength\tabcolsep{0.5pt}   \small
\begin{tabular}{@{}l||c|c@{}} 
\dtoprule 
\multicolumn{1}{c||}{Low Inter-Client}   & \multirow{2}{*}{\shortstack{CIFAR10}} & \multirow{2}{*}{\shortstack{OrganAMNIST}} \\ 
\multicolumn{1}{c||}{Data Heterogeneity} & & \\ \hline
 ~~FedAvg ($100\%$) & $57.17~{\scriptstyle (\pm 0.34)}$ & $81.49~{\scriptstyle (\pm 0.96)}$ \\  
 ~~FedProx ($100\%$) & $\mathbf{57.45}~{\scriptstyle (\pm 0.51)}$  & $\mathbf{82.07}~{\scriptstyle (\pm 1.10)}$ \\  \hdashline
~~FedAvg & $40.11~{\scriptstyle (\pm 0.67)}$ & $59.74~{\scriptstyle (\pm 1.52)}$ \\  
 ~~FedProx & $41.45~{\scriptstyle (\pm 0.58)}$ & $61.86~{\scriptstyle (\pm 1.23)}$  \\  \hdashline
~~FedAvg+UDA & $42.07~{\scriptstyle (\pm 0.94)}$ & $63.05~{\scriptstyle (\pm 1.91)}$ \\  
~~FedAvg+FixMatch & $38.51~{\scriptstyle (\pm 0.96)}$ & $70.14~{\scriptstyle (\pm 1.42)}$ \\  
~~FedProx+UDA & $42.08~{\scriptstyle (\pm 0.83)}$ & $65.12~{\scriptstyle (\pm 2.03)}$ \\  
~~FedProx+FixMatch & $42.94~{\scriptstyle (\pm 0.86)}$ & $70.58~{\scriptstyle (\pm 1.13)}$ \\  
 ~~FedTriNet & $42.67~{\scriptstyle (\pm 0.79)}$  & $74.16~{\scriptstyle (\pm 1.42)}$   \\ 
  ~~FedMatch & $44.23~{\scriptstyle (\pm 0.88)}$ & $72.57~{\scriptstyle (\pm 1.58)}$  \\ 
~~\fedl~(\textbf{ours}) & $\mathbf{48.89}~{\scriptstyle (\pm 0.91)}$ & $\mathbf{79.76}~{\scriptstyle (\pm 0.83)}$  \\
 \dbottomrule   
\end{tabular}} {\vspace{-0.5em}\caption{ Test accuracy for labeled data $20\%$ and $5\%$ respectively for CIFAR10 and OrganAMNIST with lower data heterogeneity ($\alpha=1$). \fedl~outperforms the baselines by $4$-$10\%$ and $7$-$20\%$ for CIFAR10 and OrganAMNIST respectively.  \label{tab:testacc_a1} }}
\ffigbox{\includegraphics[width=0.5\textwidth]{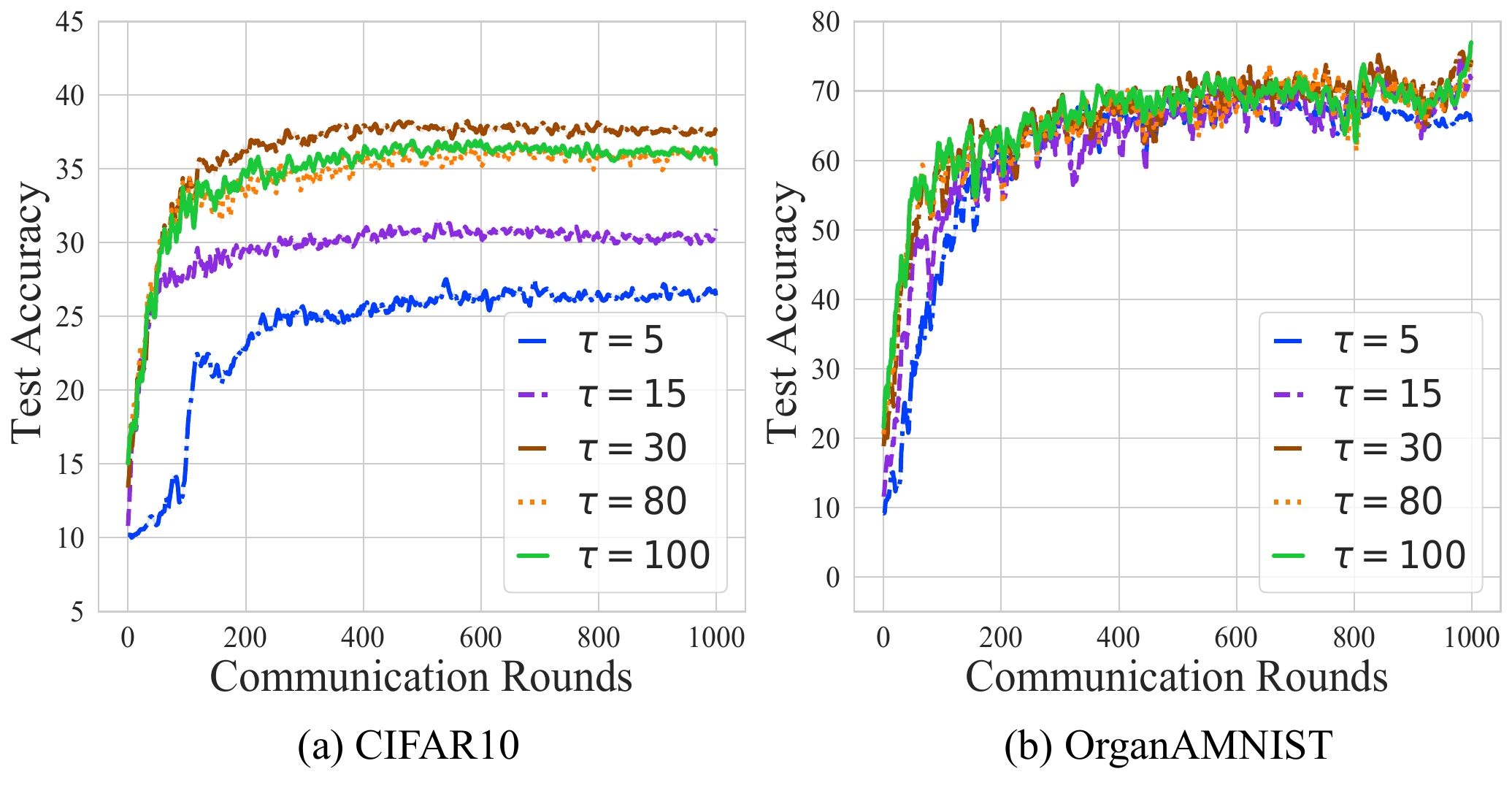}} {\caption{ Ablation study on the number of local steps $\tau$ to obtain the local model $\mathbf{w}_{\mathcal{L},k}$ for client $k\in[M]$. The smallest $\tau=5$ yields the worst performance showing that for the local model to well reflect the client's local data, $\tau$ needs to be moderately large.} 
\label{fig:abloc}  }
\end{floatrow}
\end{figure}

\noindent \textbf{Robustness of \fedl~to Data Heterogeneity.} In \Cref{tab:testacc_a1}, we show the test accuracy for smaller data heterogeneity ($\alpha=1$) where \fedl~still outperforms the other baselines by $4$-$10\%$ and $7$-$20\%$ for CIFAR10 and OrganAMNIST respectively. However, compared to the results in \Cref{tab:testacc} which is for higher data heterogeneity ($\alpha=0.1$), the performance gap between \fedl~and the other baselines is smaller by around $11$-$14\%$. This implies that \fedl~works better when there is high data heterogeneity, while the other baselines perform worse when there is higher data heterogeneity across clients. \\ \vspace{-1em}

\noindent \textbf{Effect of Global-Local Consistency Regularization.} In \fedl, we use global-local consistency regularization which is weighted by the parameter $\lambda_0$ (see \Cref{eq:loc-glob2}). 
We evaluate the effectiveness of this term by varying $\lambda_0$ in \Cref{fig:globlocabl}. For all datasets, $\lambda_0=0$ gives the lowest test accuracy, showing that without the regularizing term, we are losing useful information from the discarded model due to the binary selection between the local and global model. As we increase $\lambda_0$, we see significant improvement in the test accuracy of $5$-$9\%$. However, we also observe that when $\lambda_0$ exceeds a certain threshold the improvement decreases. The intensity of the global-local consistency regularizing term can be modulated by tuning $\lambda_0$ when appropriate. \vspace{-1em}

\paragraph{Number of Training Steps to Obtain the Local Model.} To obtain the local model (see Eq. \Cref{eqn:local_model_update}), clients perform $\tau$ local SGD steps on the received global model with their labeled data. Hence, with larger $\tau$, the more the local model well reflects the client's local data. In \Cref{fig:abloc}, we perform an ablation study on the effect of $\tau$. The smallest $\tau=5$ yields the worst performance showing that for the local model to well reflect the client's local data and bring distinct information from the global model, $\tau$ needs to be set to a moderately large value. For larger $\tau$, the performance improves and gradually saturates indicating that the local model converges. We provide additional results on the effect of $\tau$ for a larger number of labeled data in \Cref{app:exprest}, giving further insight into how the quality of the local model affects \fedl's performance.

\setlength{\textfloatsep}{10pt plus 1.0pt minus 2.0pt}
\begin{wrapfigure}[10]{r}{0.5\textwidth} 
\centering \vspace{-1.5em}
\includegraphics[width=1\textwidth]{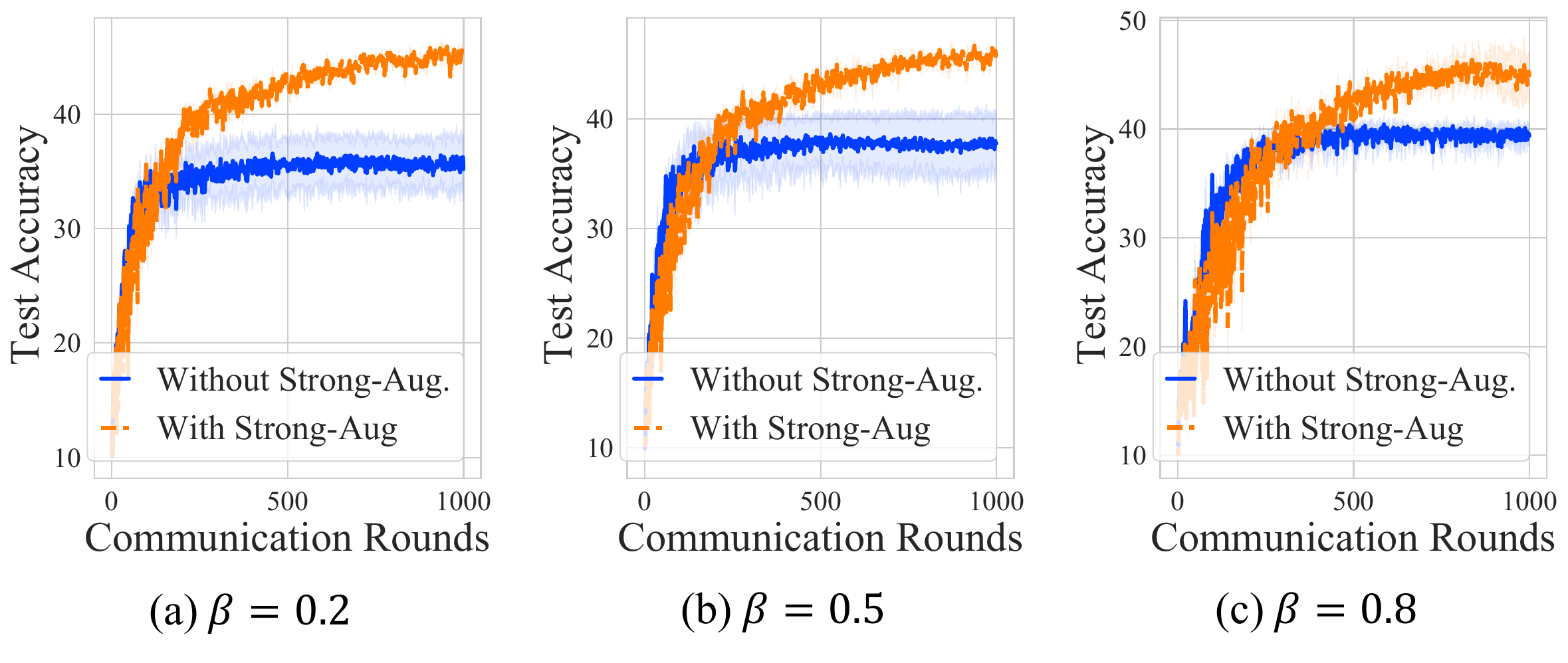} 
 \vspace{-1.5em}  \caption{\fedl~with or without strong-augumentation for CIFAR10 with varying threshold values $\beta$.}\vspace{-1em} \label{fig:aug}  
\end{wrapfigure}
\paragraph{Effect of Strong-Augumentation.} In \fedl's augumentation process, we use RandAug. [3] which randomly selects 10 of the following transformations: ['Identity', 'AutoContrast’, ’Equalize’,
’Rotate’, ’Solarize’, ’Color’, ’Posterize’,
’Contrast’, ’Brightness’, ’Sharpness’,
’Shear-X’, ’-Y’, ’Translate-X’, or ’-Y'] and applies them with magnitude 1. This acts as a regularizer, preventing the model from overfitting to the fixed set of unlabeled data, as also shown in \Cref{fig:aug} for different threshold $\beta$ values. \vspace{2em}

\renewcommand{\arraystretch}{1.2}
\begin{wraptable}[9]{!r}{0.6\textwidth} \centering \small
\setlength\tabcolsep{1.1pt} \vspace{-1em}
\begin{tabular}{@{}l||c|c|c@{}} 
\dtoprule 
{(Per comm. round \& client)} &  FedTriNet & FedMatch & \fedl \\ \hline
Forward+Backward Pass & $3P+2B$ & $(h+1)P+2B$ & $2P+2B$\\  
Additonal Server Comp. & Thresholding & Clustering ($\propto h$) & None  \\ \hdashline
Comm. Cost & $2R$ & $(2+h)R$ & $2R$ \\
Comp. Time (sec)& $7.81\scriptstyle (\pm 0.16)$ & $11.24\scriptstyle (\pm 0.21)$ &$3.76\scriptstyle (\pm 0.04)$ \\
 \dbottomrule   
\end{tabular} \vspace{-0.8em} \caption{Computation \& communication cost comparison for different semi-supervised FL methods. \label{tab:comp-cost} }
\end{wraptable} \vspace{-0.8em}
\paragraph{Computation Cost of \fedl.} We further investigate \fedl's computation efficacy in \Cref{tab:comp-cost}. Assuming the same number of local updates and batch sizes across all baselines for fair comparison, and setting the \# of model params as $R$, \# of helper clients as $h=2>1$ for FedMatch, computation cost for a model's single forward \& backprop. as $P$ \& $B$. The actual time taken per computation step for each client plus any additional computations taken at the server side is  presented where \fedl~incurs significantly smaller computation time. Costs for sending constants such as the max-probability for FedTriNet, and weights for \fedl~\& FedMatch are considered negligible.

\vspace{-0.5em}
\section{Concluding Remarks} \vspace{-0.5em}
In conclusion, we propose \fedl, a SSFL framework that works well for both low and high data heterogeneity cases, as well as for both limited and larger portions of labeled data. We use a confidence-based binary selection of the local or global model for pseudo-labeling with global-local consistency regularization. Unlike previous work, \fedl~does not require additional computation to find new experts, additional communication of parameters, server labeled data, or any fully labeled clients. In both cross-device and cross-silo settings, we show that \fedl~largely outperforms other SSFL baselines, especially when there is high data heterogeneity and label scarcity, by at most $24\%$. \fedl~even outperforms fully-supervised FL baselines which use fully-labeled data with only using $5$-$20\%$ of labeled data. Currently, \fedl~does not consider the possible noise that can be present in the labeled data of the clients caused, for instance, clients mislabeling their data due to lack of expertise. Thus, for future work, we aim to extend the \fedl~to be robust to label noise.

\newpage
{
\bibliography{dist_sgd}
\bibliographystyle{unsrt}

 \newpage
\appendix
\section{Additional Experimental Results} 
\begin{figure*}[!h]
\centering
\begin{subfigure}{0.24\textwidth}
\centering
\includegraphics[width=1\textwidth]{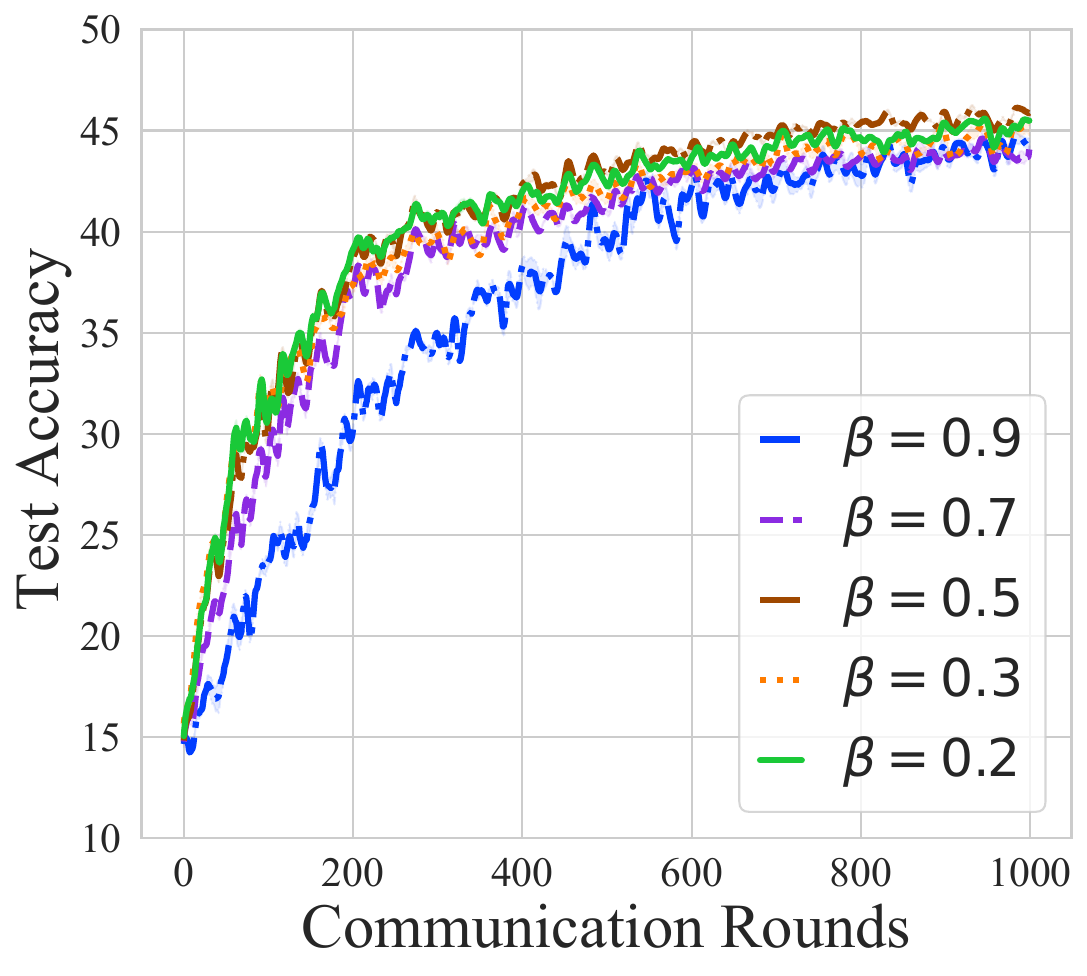} \caption{ CIFAR10}
\end{subfigure} \hfill
\begin{subfigure}{0.24\textwidth}
\centering
\includegraphics[width=1\textwidth]{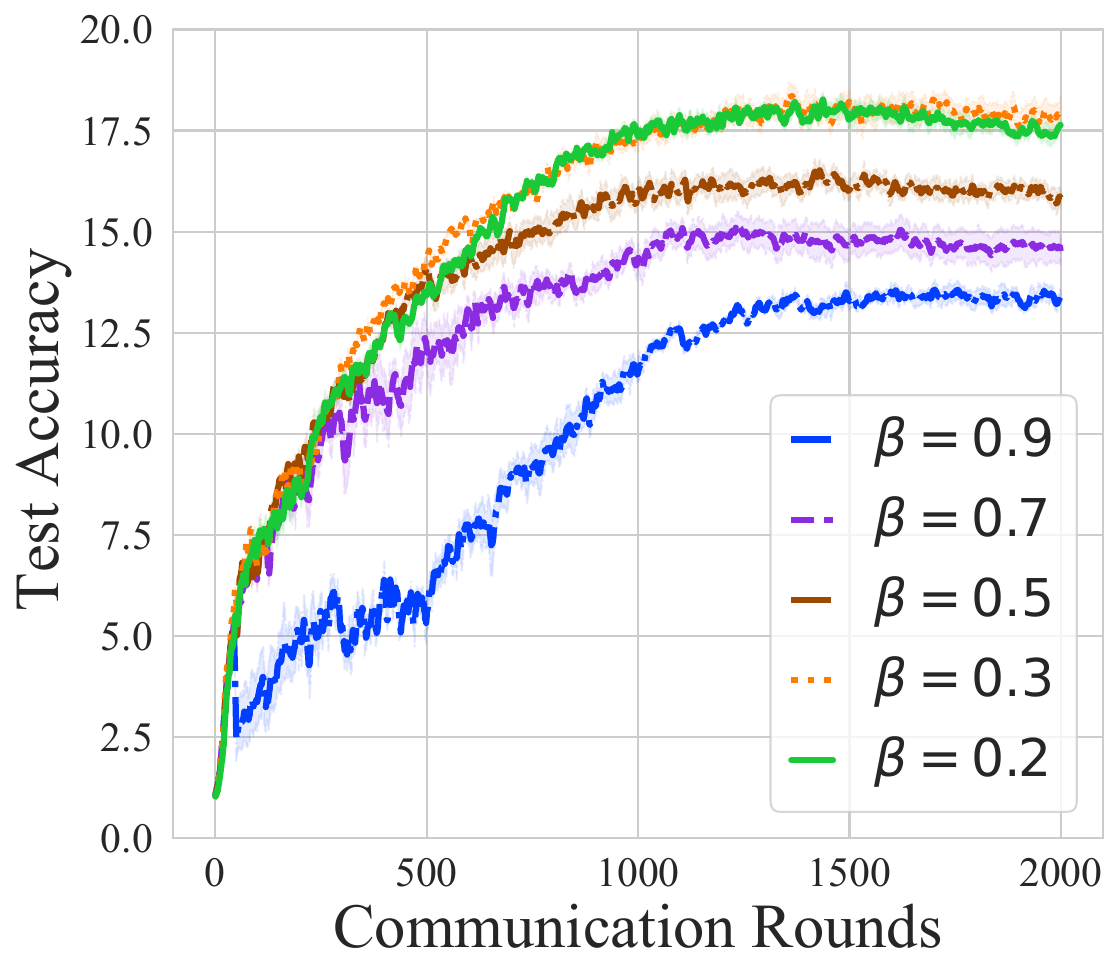} \caption{ CIFAR100}
\end{subfigure} \hfill
\begin{subfigure}{0.24\textwidth}
\centering
\includegraphics[width=1\textwidth]{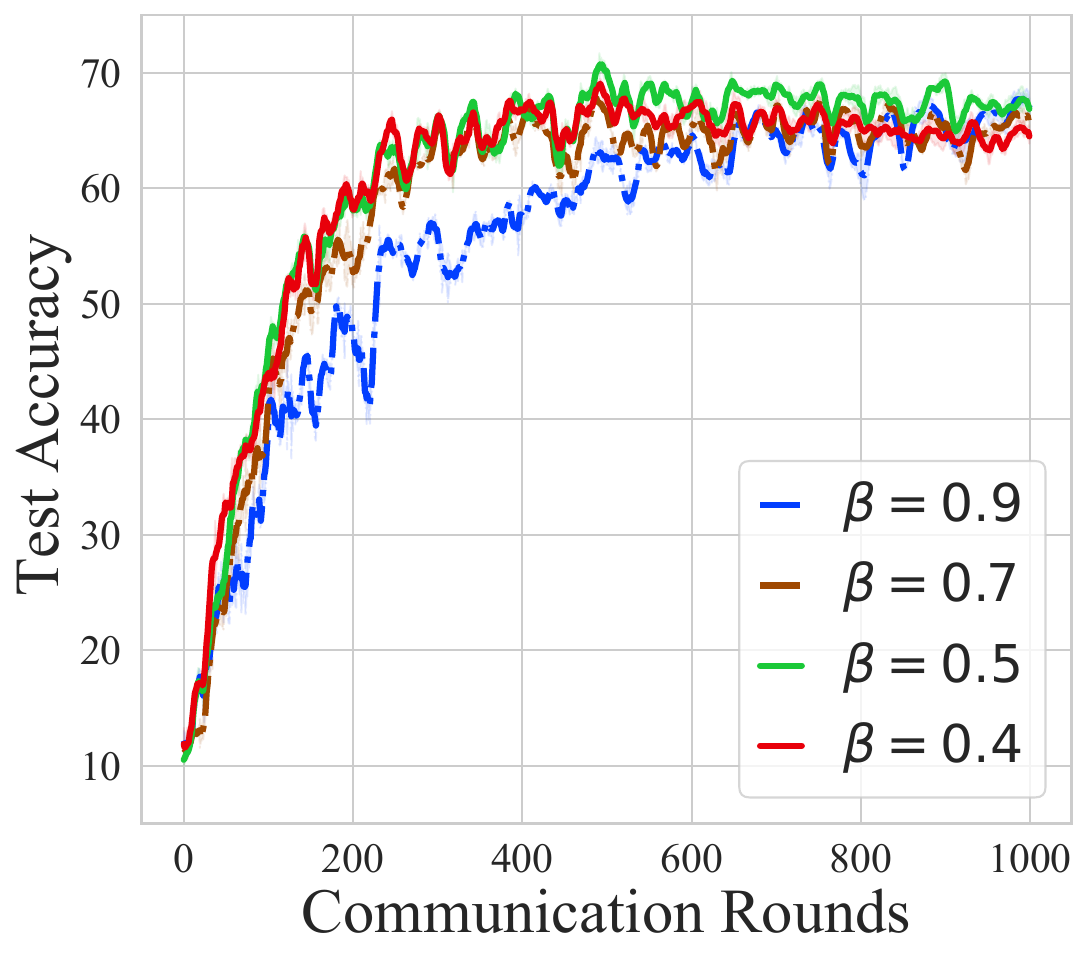} \caption{ OrganAMNIST}
\end{subfigure} \hfill
\begin{subfigure}{0.24\textwidth}
\centering
\includegraphics[width=1\textwidth]{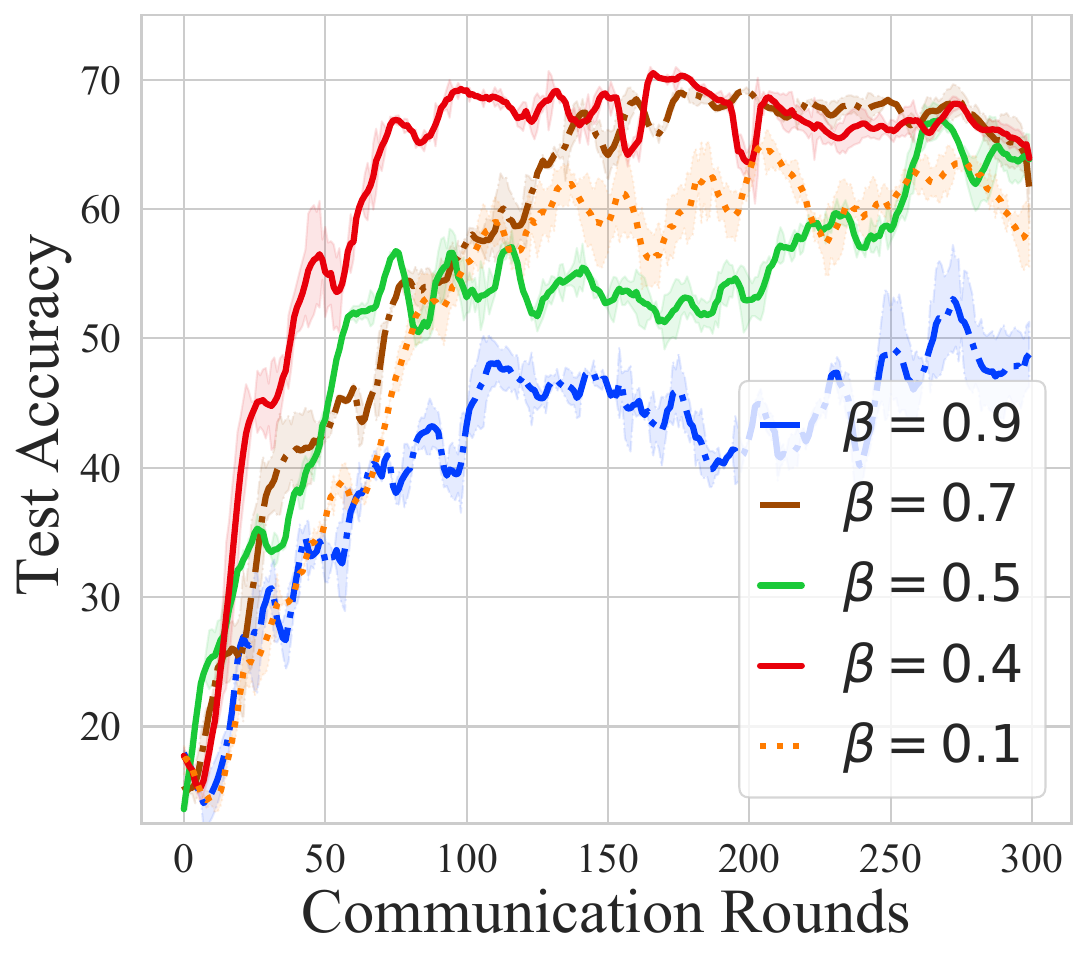} \caption{ BloodMNIST}
\end{subfigure} \hfill 
 \caption{ Ablation study of \fedl's performance on the thresholding parameter ($0\leq\beta\leq 1$ in \Cref{eq:pseul}). For all the datasets, the performance of \fedl~is the lowest for the highest $\beta=0.9$ meaning that for a too high $\beta$ parameter, \fedl~filters out too much unlabeled data that the model is not able to learn much. \fedl~achieves the best performance for a lower $\beta$ of the range $0.3\sim 0.5$.}
\label{fig:threshold} 
\end{figure*}

\begin{figure*}[!h]
\centering
\begin{subfigure}{0.24\textwidth}
\centering
\includegraphics[width=1\textwidth]{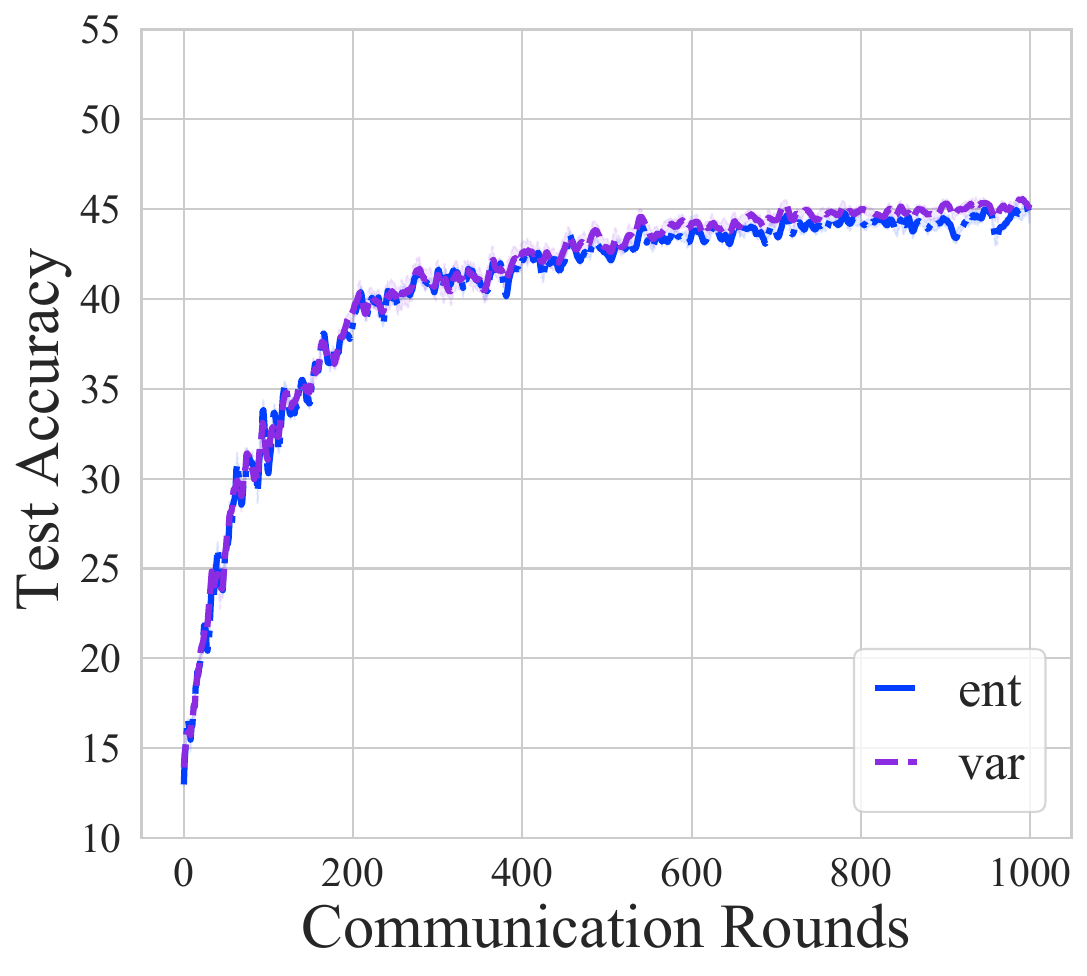} \caption{ CIFAR10}
\end{subfigure} \hfill
\begin{subfigure}{0.24\textwidth}
\centering
\includegraphics[width=1\textwidth]{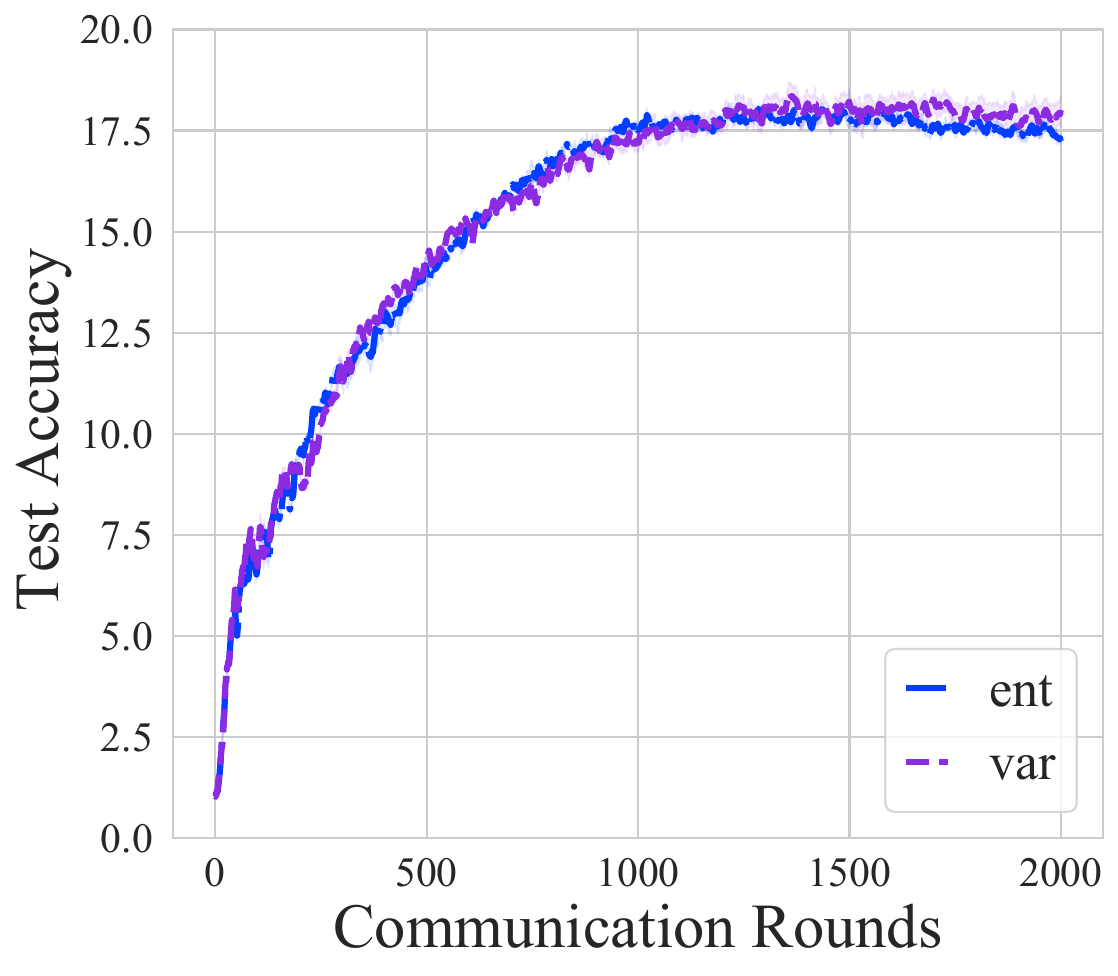} \caption{ CIFAR100}
\end{subfigure} \hfill
\begin{subfigure}{0.24\textwidth}
\centering
\includegraphics[width=1\textwidth]{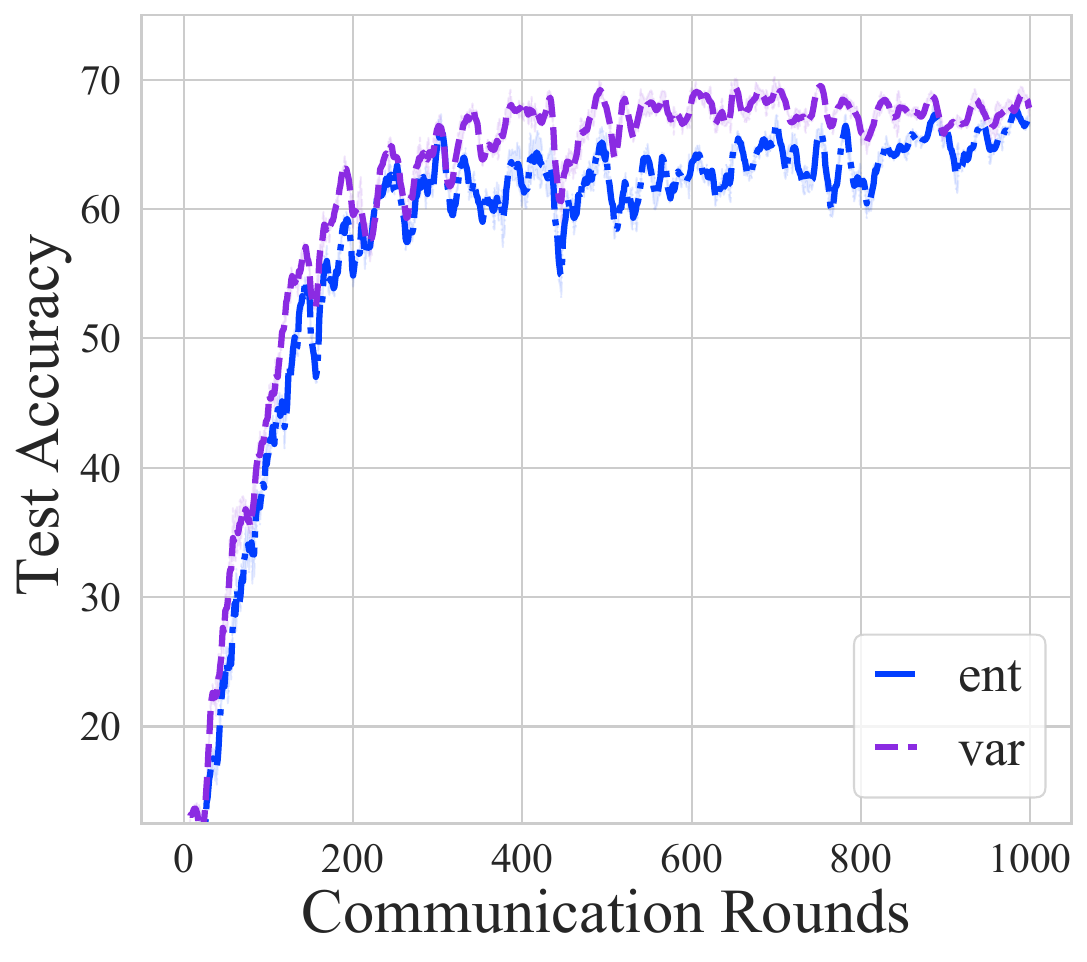} \caption{ OrganAMNIST}
\end{subfigure} \hfill
\begin{subfigure}{0.24\textwidth}
\centering
\includegraphics[width=1\textwidth]{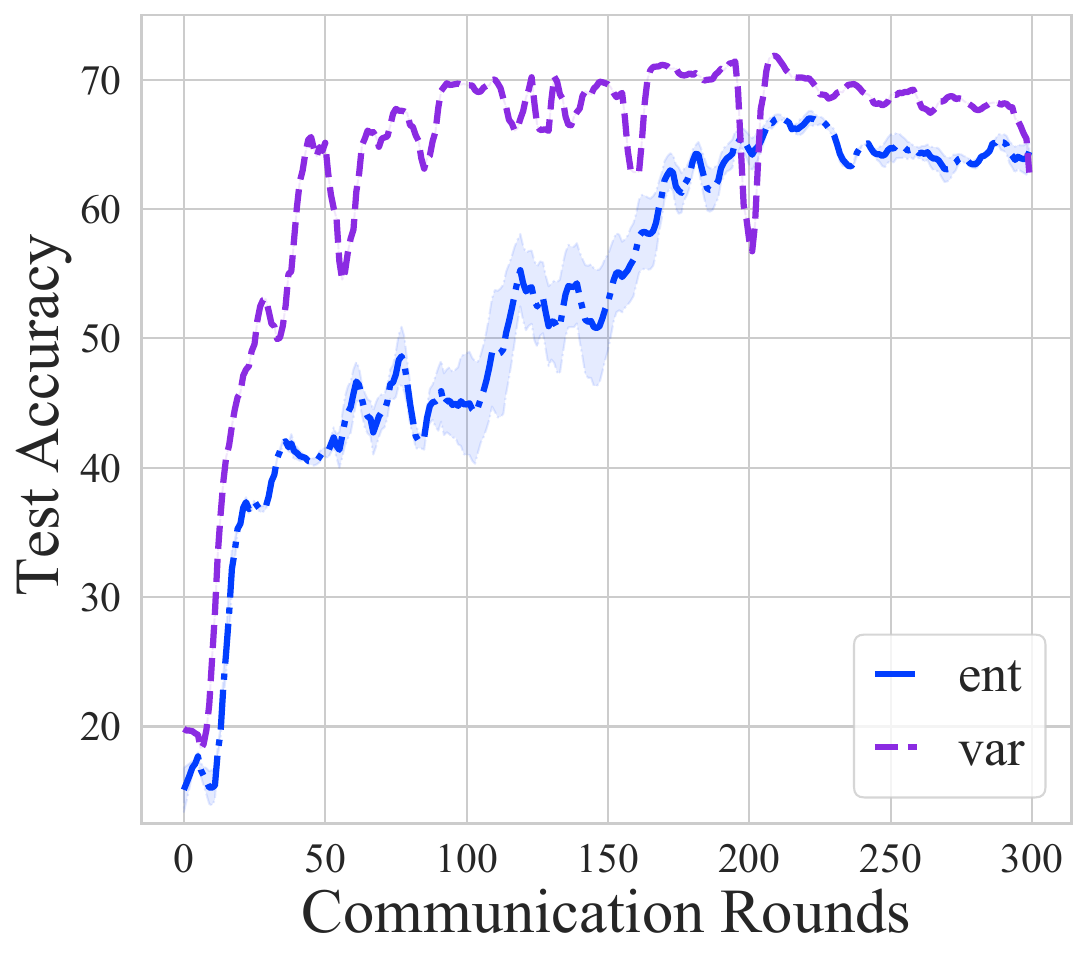} \caption{ BloodMNIST}
\end{subfigure} \hfill 
 \caption{ \fedl's performance with different confidence metrics ($h(\cdot)$ in \Cref{eq:pseul}), variance and entropy. For CIFAR10 and CIFAR100 both entropy and variance perform similarly. However, for OrganAMNIST and BloodMNIST variance performs better than entropy showing that it can better judge the confidence of the models than entropy.}
\label{fig:conf} 
\end{figure*}

For all datasets, $80\%$ is for training partitioned across clients and the $5\%,~15\%$ of the data is for the validation and test respectively. We experiment with 3 different seeds for the randomness in the dataset partition across clients and present the averaged results. \\  \label{app:exprest}

\noindent \textbf{Ablation Study on Thresholding.} The thresholding parameter of \fedl~($0\leq\beta\leq1$ in \Cref{eq:pseul}) determines whether either the local or global model produces a high-enough confidence logit to be used for training which has been proven to be effective for SSL~\cite{sohn2020fixmatch}. We perform an ablation study on the different values of $0\leq\beta\leq1$ and its effect on the overall performance of \fedl~in \Cref{fig:threshold}. In \Cref{fig:threshold}, for all different datasets for a high $\beta=0.9$, the test accuracy is the lowest, implying that if we set $\beta$ to a too high value, \fedl~filters out too much unlabeled data and the model is not able to learn effectively. As we lower the threshold $\beta<0.9$, the test accuracy improves significantly by at most approximately $17\%$. We observe that the best performing $\beta$ value can be surprisingly low to around the range of $\beta\in[0.3,0.5]$ for the different datasets. This is due to the RandAugumentation step which heavily transforms the image so that the image becomes different from the images the clients have been training. Hence the overall confidence on the strongly-augumented image becomes lower as observed in \Cref{fig:threshold}.

\noindent \textbf{Gauaging Confidence of the Models.}
One may wonder what is the appropriate metric to measure the confidence of the models' logits ($h(\cdot)$ in \Cref{eq:pseul0}). While we use variance as the confidence metric for all the other results in our work, we investigate how another representative measure for confidence, entropy, compares to the variance metric. Entropy is a commonly used metric to measure the uncertainty of a probability distribution, and since logits are discrete probabilities it can also be presented as an adequate candidate for measuring the confidence of the models. In \Cref{fig:conf}, we show that for CIFAR10 and CIFAR100 both entropy and variance performs similarly. However, for OrganAMNIST and BloodMNIST variance performs better than entropy showing that it can better judge the confidence of the models than entropy. We conjecture that this is due to entropy compressing the confidence values to be between $[0,1]$ while variance doesn't have this, allowing to compare the confidence of the different logits more accurately. \\

\begin{wrapfigure}{r}{0.5\textwidth} 
\centering \vspace{-1.5em}
\includegraphics[width=1\textwidth]{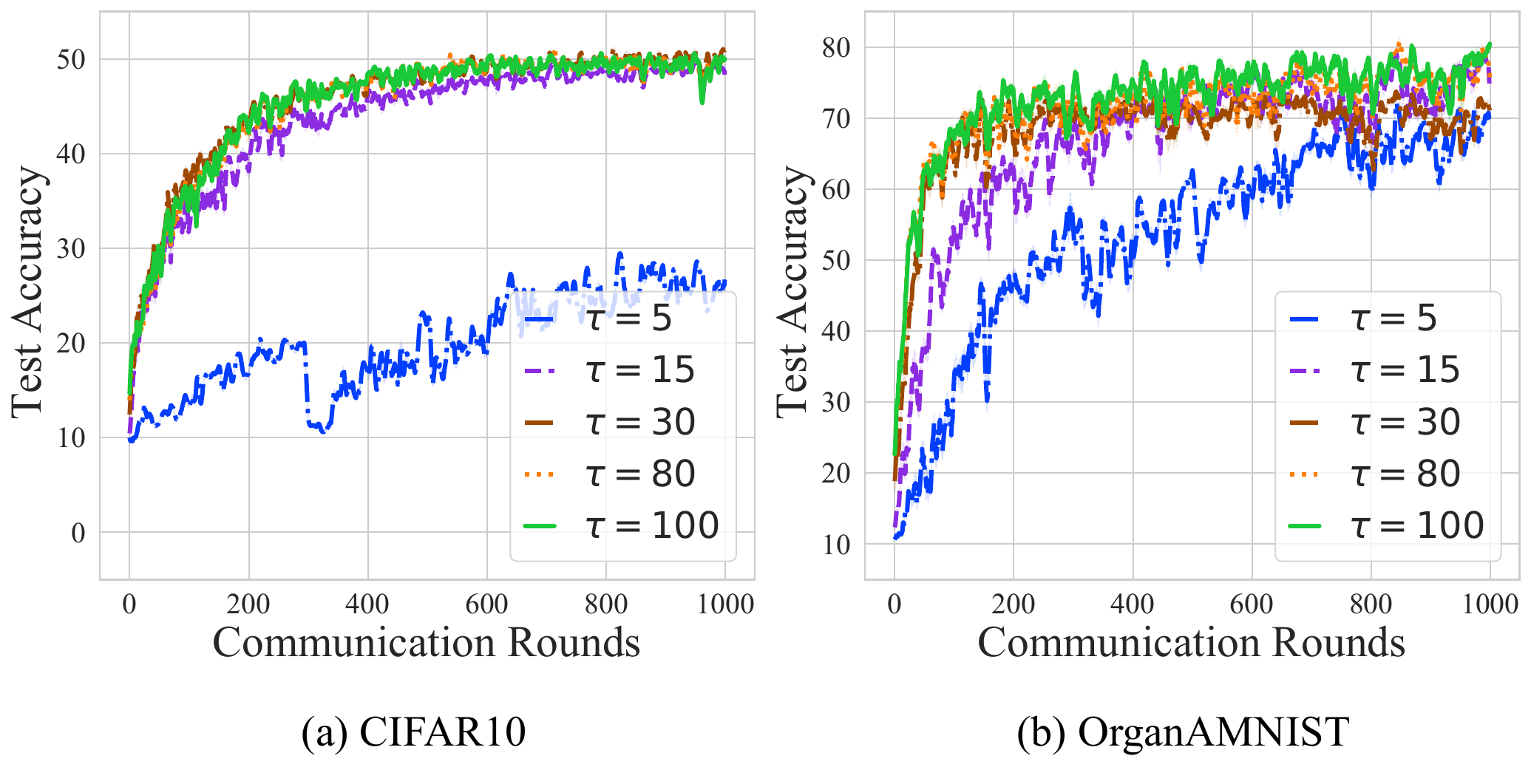} \caption{ Ablation study on the number of local steps $\tau$ to obtain the local model $\mathbf{w}_{\mathcal{L},k}$ for client $k\in[M]$ for larger number of labeled data ($50\%$ for CIFAR10 and $20\%$ for OrganAMNIST). The smallest $\tau=5$ yields the worst performance, but a slightly larger $\tau$ can largely improve the performance.} \vspace{-1em}
\label{fig:abloc_50}  
\end{wrapfigure}
\noindent \textbf{Number of Training Steps to Obtain the Local Model for Larger Number of Labeled Data.} For a larger number of labeled data, compared to the results in \Cref{fig:abloc}, we see in \Cref{fig:abloc_50} that the smallest $\tau=5$ also yields the worst performance across the range of $\tau$. However, even for a slightly larger $\tau$, we can see that the performance improves with a much larger gap than the results in \Cref{fig:abloc}. This indicates that with a larger number of labeled data, even a slightly large $\tau$ can make the local model sufficiently represent the local data of the client, and improve the performance of \fedl.

\end{document}